\documentclass[letterpaper, 10 pt, journal]{IEEEtran}
\usepackage{cite}
\usepackage{graphicx} 
\usepackage{stfloats}
\usepackage{url}
\usepackage[dvipsnames]{xcolor}
\usepackage{epstopdf}
\graphicspath{./images}
\DeclareGraphicsExtensions{.pdf,.jpeg,.png}
\usepackage{multicol}
\usepackage[left=0.65in,top=0.7in,right=0.65in,bottom=0.7in]{geometry}

\begin{document}
\title{A Comprehensive Review of Smart Wheelchairs: Past, Present and Future}
\author{Jesse Leaman, and Hung M. La, \textit{Senior Member, IEEE }
\vspace{-20pt}
\thanks{The authors are with  the Advanced Robotics and Automation (ARA) Lab, Department of Computer Science and Engineering, University of Nevada, Reno, NV 89557, USA. Corresponding author: Hung La, email: hla@unr.edu}
}

\vspace{-5pt}
\maketitle

\begin{abstract}
A smart wheelchair (SW) is a power wheelchair (PW) to which computers, sensors, and assistive technology are attached. In the past decade, there has been little effort to provide a systematic review of SW research. This paper aims to provide a complete state-of-the-art overview of SW research trends.  We expect that the information gathered in this study will enhance awareness of the status of contemporary PW as well as SW technology, and increase the functional mobility of people who use PWs. We systematically present the international SW research effort, starting with an introduction to power wheelchairs and the communities they serve. Then we discuss in detail the SW and associated technological innovations with an emphasis on the most researched areas, generating the most interest for future research and development. We conclude with our vision for the future of SW research and how to best serve people with all types of disabilities.
\end{abstract}


\section{Introduction and Review Methodology}
\IEEEPARstart{P}{eople} with cognitive/motor/sensory impairment, whether it is due to disability or disease, rely on power wheelchairs (PW) for their mobility needs. Since some people with disabilities cannot use a traditional joystick to navigate their PW they use alternative control systems like head joysticks, chin joysticks, sip-n-puff, and thought control \cite{Mazumder2014, pasteau2014, Desmond2013_TePRA, Sinyukov2014, Millan2014_BIC}. In many cases PW users have difficulties with daily maneuvering tasks and would benefit from an automated navigation system. Mobility aside, people with disabilities are heavily reliant on their caregivers for eating and drinking, handling items, and communicating with others, especially in large groups.

To accommodate the population of individuals who find it difficult or impossible to operate a PW, several researchers have used technologies originally developed for mobile robots to create  smart wheelchairs \cite{Rathore2014, Yayan2014, Sinyukov2014, Leishman2014, Jain2014_IROS}. A smart wheelchair (SW) typically consists of either a standard PW base to which a computer and a collection of sensors have been added, or a mobile robot base to which a seat has been attached \cite{Simpson:2004}. Pineau et al. 2011 argue that the transition to wheelchairs that cooperate with the user is at least as important as that from manual to powered wheelchairs, possibly even more important since this would mark a paradigmatic rather than merely a technological shift \cite{Pineau:2011}. 

Overall, there has been a huge effort in research and development (R\&D) on both PWs and SWs in particularly, and assistive technology in general for people with disabilities. There have been many reported R\&D successes, like innovative autonomous SWs \cite{Chao2008_JFR, Andrew2010_FSR, Li2013}. However, in the past decade since the work by Simpson \cite{Simpson2005} and Ding et al. in 2005 \cite{Ding2005_CS}, there has been little effort to extensively and intensively review, and more importantly provide a complete state-of-the-art overview of the research trends for SWs. To fill this gap, in this paper we aim to provide the readers and researchers a systematic and comprehensive review of the international SW research efforts since the major review works by Simpson and Ding in 2005 \cite{Simpson2005, Ding2005_CS}. 

\begin{figure*}[t!]
\centering
\includegraphics[width=\textwidth]{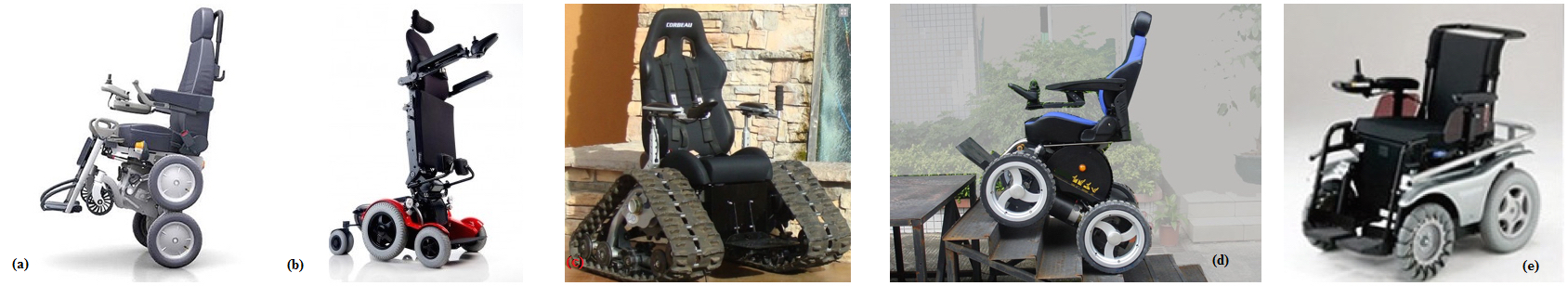}
\caption{Examples of modern PW form factors: (a) stair climbing iBot [discontinued 2008] \cite{iBot}, (b) Australia's Ability in Motion \cite{abilityinmotion}, standing wheelchair, (c) the Tank Chair \cite{tankchair}, (d) Chinese Observer remains level on all-terrain \cite{observer}, and (e) Patrafour by Toyota Motor East Japan  \cite{Patrafour}.}
\label{forms}
\vspace{-10pt}
\end{figure*}

To that end we conducted a literature search in major electronic databases including, IEEE Xplore, Google Scholar and PubMed. The articles were included in this systematic review when they met the following criteria: (i) they were written in English; (ii) they were published from 2005 to 2015; and (iii) the conference proceedings were also examined only if they were substantially different from journal articles presented by the same authors. Titles and abstracts were first screened, and then the full text copies of all studies were examined to determine whether they met the pre-specified inclusion criteria. The essential component of our search keyword should be \emph{``wheelchair''} then we add more supplemental words to form our keywords. We started with \emph{powered wheelchair} and \emph{electric wheelchair} keywords to acquire an overall result. Since the search produced  a significantly large number of results, we refined the keywords by grouping or adding more specific keywords such as \emph{``powered wheelchair''}, \emph{``electric wheelchair''}, \emph{``intelligent wheelchair"}, \emph{``autonomous wheelchair"}, \emph{``robotic wheelchair"}, and \emph{[``human factors" + ``wheelchair'']} to narrow down the search area and increase the chance of getting relevant hits. We also create several combinations using those keywords to enhance our search. Table \ref{tab:lsr} illustrates the search results when using a set of keywords on different popular databases. At the end, the search revealed 155 
 references that described achievements in SW research, which we present in the chronology shown in Table \ref{tab:gl}.

\begin{table}[h]
\caption{Example literature search results on several electronic databases.}
\label{tab:lsr}
\centering
\begin{tabular}{lllc}
\hline
Keywords  & Google Scholar & IEEE Xplore   \\
\hline
 Powered wheelchair  & 17500 & 487  \\
 Electric wheelchair  & 16900 & 367  \\
 ``Powered wheelchair''  & 4450 & 140  \\
 ``Electric wheelchair'' & 5,260 & 160 \\
 ``Intelligent wheelchair'' & 2480 & 131\\
 ``Autonomous wheelchair'' & 952 & 35\\
 ``Robotic wheelchair'' & 2300 & 131 \\
 ``Human factors'' + ``wheelchair'' & 715 & 1\\
\hline
\end{tabular}
\vspace{-5pt}
\end{table}

\begin{table}[h]
\caption{Our literature search revealed 155 references divided into: past research on which present projects rely; present research that is guided by human factors to produce the best solutions to current challenges; and research on SW features that will be part of future applications.}
\label{tab:gl}
\centering
\begin{tabular}{llc}
\hline
& Subcategory & Count \\
\hline
Past & & \\
& Haptic feedback & 3 \\
& Voice recognition  & 5 \\
& Object recognition  & 3 \\
& Human physiology & 4  \\
& Collision avoidance & 8 \\
& Social issues & 12  \\
\hline
Present  & & \\
& Cloud computing  & 3 \\
& Computer vision  & 8 \\
& Game controller  & 3 \\
& Touch screens & 5 \\
& Following  & 10 \\
& Machine learning  & 19 \\
& Mapping & 3 \\
& Navigational assistance  & 25\\
& Human learning  & 9   \\
& Operations  & 11    \\
\hline
Future  &  \\
& Biometrics  & 3  \\
& Brain Computer Interface & 11 \\
& Multi-modal input methods & 3 \\
& Localization & 2 \\
& Human Robot Interaction  & 5  \\
\hline
 Total & & 155 \\
 \hline
\end{tabular}
\vspace{-5pt}
\end{table}

 We review the concepts and showcase the latest human-computer interface hardware, sensor processing algorithms, and machine-vision innovations made in recent years. These tools give people with disabilities not only mobility but also the necessary help and support to handle daily living activities.  We hope that the information gathered in this study will enhance awareness of the status of contemporary SW technology, and ultimately increase the functional mobility and productivity of people who use PWs.

The rest of this paper is organized as follows: next we present past achievements in PW development (Sec. \ref{Past}), followed by a description of the international SW research effort and an overview of the main topics of interest (Sec. \ref{Current}). Then we showcase the particular research in the areas of input methods (Sec. \ref{im}), operating modes (Sec. \ref{om}), and human factors (Sec. \ref{hf}). Finally, we lay out our vision of the future of SW research and development (Sec. \ref{Future}), and conclude with a discussion of changes since the last review (Sec. \ref{Con}). 

\vspace{-5pt}
\section{Past: The Power Wheelchair}\label{Past}
George Klein invented the first PW for people with quadriplegia injured in world war II while he was working as a mechanical engineer for the National Research Council of Canada \cite{Globe2012}. A person with that level of disability is by default bedridden because without assistance from people and/or technology there would be no way to get up. That being said, with the right combination of human and technological resources, a person with quadriplegia can still have an extremely rich life and make a major contribution to society, a philosophy we hope to share with people who have just experienced the onset of disability.

While primarily intended for people with mobility disabilities, those suffering from ailments resulting in fatigue, and pain can also benefit from a PW. By 1956 Everest \& Jennings and the American Wheelchair Company began producing PWs for mass sales \cite{EJ1956}. The basic components of a PW are: 

\begin{itemize}
\item{Chassis, or drive system, may be front-wheel, rear-wheel, center-wheel, or all-wheel-drive. Chassis may be foldable \cite{Smartchair}, include stairclimbing ability (Fig. \ref{forms}a), have standing ability (Fig. \ref{forms}b), have all-terrain tank tracks (Fig. \ref{forms}c), or 4-wheel all-terrain drive (Fig. \ref{forms}d, \ref{forms}e).
}
\item{Batteries: The first PWs derived their power from 2- 24V wet cell batteries. But these batteries have to be removed from the wheelchair during travel on airplanes. They were eventually replaced by dry cell batteries.
}
\item{Controller is the interface between human and machine. Commercially available controllers include: hand joystick, sip-n-puff, chin joystick, and head joystick.
}
\item{Seating system: Seats, are typically upgraded to include cushions that use foam, gel, or air to prevent pressure sores; Backrests, are typically padded with foam and can be motorized to tilt and recline; Lateral supports keep the user from tilting side-to-side; and Footrests, are either removable or motorized to accommodate a more comfortable reclining position.
}
\end{itemize}

Medicare prescription guidelines limit PWs to those individuals who cannot use a manual wheelchair. But many more people would benefit from a PW, and often resort to less expensive, generic solutions that often do not meet the needs of the individual and their particular disability. 

Some people with quadriplegia end up making major modifications to their own wheelchairs in order to have additional safety measures, like lights or reflectors, rearview cameras, and assistive technology to use as an input method for computing.

Co-author, Dr. Jesse Leaman, began his quest to improve the PW user experience in 1998 while a summer intern at NASA's Marshall Space Flight Center. By 2007, his invention, the information technology upgrade package for PWs, dubbed ``Gryphon Shield", was recognized as one of the the year's top 25 inventions by the History Channel and the National Inventors Hall of Fame \cite{LeamanNews}. By 2010, however the system had grown too large, and heavy for the wheelchair base, and had become difficult to maintain. The upgrade of ``Gryphon Shield'' is the iChair presented in \cite{iChair, Leaman_MFI16} where we take advantage of a head tracking mouse \cite{Headmouse} and a Mount-n-Mover \cite{MountnMover}.

Overall, past studies have successfully  achieved initial technological advancements that aid PW users in their daily activities. Nevertheless, there is little assistive technology included in the wheelchair to make it ``smart,'' which is the goal of  research on integrating ``intelligent'' technology on a PW, called ``smart wheelchair''.

\vspace{-5pt}
\section{Present: The Smart Wheelchair}
\label{Current}
This section presents current SW research achievements around the world  revealing a major shift from PW to SW.

Fig. \ref{forms} shows several recent power wheelchair options that can be used as the platform for a SW. Some early SWs were mobile robots with a seat attached \cite{VAHM, MrED}, but most were based on heavily modified commercially available PWs to which computers and sensors are attached \cite{Simpson2005}.  The majority of SWs that have been developed to date have been tightly integrated with the underlying PW, requiring significant modifications to function properly \cite{Simpson:2004, Simpson:2008}. Ideally the SW system should easily be removable from the underlying wheelchair so that the user can attach it to a different chair, which is especially important for children, who may go through several wheelchairs before they are all grown-up \cite{Pineau:2011}.

\begin{table}[h!]
\caption{SW research has become a truly international effort. Below is a list of Institutions around the world that have produced a SW prototype since 2005. For a review of research efforts prior to 2005 see S05 \cite{Simpson2005}.}
\label{tab:overview}
\centering
\begin{tabular}{lcc}
\hline
Description & Location & Year \\
\hline
US Naval Academy \cite{Navysmartchair} & Maryland, USA & 2005 \\
Lehigh University \cite{Gao2007} & Pennsylvania, USA & 2007 \\
University of Pittsburgh \cite{Ding2007} & Pennsylvania, USA & 2007 \\
Imperial College \cite{Carlson2008,Carlson2012} & London, UK & 2008 \\
Doshisha University \cite{Miyazaki2009} & Kyoto, Japan & 2009 \\
University of Bremen \cite{Rolland} & Bremen, Germany & 2009 \\
University of Paris 8 \cite{Touati2009} & Paris, France & 2009 \\
Chuo University \cite{Niitsuma2011} & Tokyo, Japan & 2011 \\
Toyohashi U.of Technology \cite{Veno2011} & Tokyo, Japan & 2011 \\
McGill University \cite{Pineau:2011} & Montreal, Canada & 2011 \\
University of BC \cite{Viswanathan2011, Giesbrecht2015} & Vancouver, Canada & 2011 \\
University of Aveiro \cite{Urbano2011} & Portugal & 2011 \\
Universidad Carlos III \cite{sultan2011} & Madrid, Spain & 2011 \\
KU Leuven \cite{Poorten2012} & Heverlee, Belgium & 2012 \\
University of Canterbury \cite{Tang2012} & Christchurch, NZ & 2012 \\ 
U. of Apl. Sci. W. Switzerland \cite{Carrino2012} & Fribourg, Switzerland & 2012 \\
University of Sydney \cite{Nguyen2012} & Sydney, Australia & 2012 \\
Case Western Reserve \cite{Cockrell2013} & Ohio, USA & 2013 \\
Hefestos \cite{Tavares2013} & Sao Leopoldo, Brazil & 2013 \\
Indian Institute of Technology \cite{Trivedi2013} & Jodhpur, India & 2013 \\
Institute of Engg. \& Technology \cite{Tyagi2013} & Ghaziabad, India & 2013 \\
ATRII \cite{Morales2013} & Kansai, Japan & 2013 \\
Chonnam National University \cite{Park2013} & Gwangju, S. Korea & 2013 \\
King Abdulaziz University \cite{Fattouh2013} & Jeddah, Saudi Arabia & 2013 \\
U. of Alabama, Huntsville \cite{Milenkovic2013} & Alabama, USA & 2013 \\
U. of Texas, Arlington \cite{McMurrough2013} & Texas, USA & 2013 \\
B.M.S College of Engineering \cite{Akash2014} & Bangalore, India & 2014 \\
Kumamoto University \cite{SUGANO2014} & Kumamoto, Japan & 2014 \\
Saitama University \cite{Sato2014} & Saitama, Japan & 2014 \\
LURCH \cite{Senthilkumar2014} & Chennai, India & 2014 \\
Integral Rehabilitation Center \cite{CRIO2014} & Orizaba, Mexico & 2014 \\
University of Kent \cite{Henderson2014} & Canterbury UK & 2014 \\
University of Lorraine \cite{Leishman2014} & Metz, France & 2014 \\
Uni. Politecnica delle Marche \cite{Cavanini2014} & Ancona, Italy & 2014 \\
University Tun Hussein Onn \cite{Tomari2014} & Malaysia & 2014 \\
Northwestern University \cite{Jain2014} & Illinois, USA & 2014 \\
UMBC \cite{Carrington2014} & Maryland, USA & 2014 \\
Smile Rehab \cite{SRsmartchair} & Greenham, UK & 2015 \\
University of Nevada, Reno\cite{iChair} & Nevada, USA & 2015 \\
\hline
\end{tabular}
\vspace{-10pt}
\end{table}

It is challenging to build an efficient SW that people with all types of disabilities feel comfortable using. The system should be mountable on any make of PW, easily removable for maintenance and travel. Table \ref{tab:overview} lists institutions around the world that have produced a SW prototype since 2005, which illustrates that interest in the topic has grown in recent years. Sensor/software packages, like collision avoidance, have had widespread success in the automotive industry, becoming less expensive and more trusted than ever before. Based on our search results, we have divided present research into three main topics: (i) input methods; (ii) operating modes; and (iii) human factors.

\begin{table}[h]
\caption{Input methods}
\label{tab:im}
\centering
\begin{tabular}{lc}
\hline
Method  & References    \\
\hline
 Biometrics  & \cite{Fernandez-Carmona2009, Felzer2009, Yokota2011}  \\
   Brain computer interface (BCI) & \cite{Blatt2008, Shen2011, Carlson2013, Diez2013, Fattouh2013, Kannan2013, Lopes2013, Ahire2014, Champaty2014, Gandhi2014, Kaufmann2014, Senthilkumar2014, Li2015}  \\
  Cloud & \cite{Touati2009, Milenkovi2013, Park2013, Akash2014}  \\
  Computer vision & \cite{Villalta2006, Bailey2007, Ju2009, Ahmed2010, Lee2012, Megalingam2012, Perez2012, Kawarazaki2013, Rivera2013, Kawarazaki2014}  \\
Game controller  & \cite{Ashraf2011, Ashraf2011-2, Gerling2013}  \\
 Haptic feedback & \cite{Sahnoun2006, Christensen2011, Abdelkader2012, Poorten2012, Morre2015}  \\ 
 Multimodal & \cite{McMurrough2013, Serranoa2013, Srivastava2014, Reis2015}  \\
  Touch  & \cite{Tavares2013}  \\
 Voice  & \cite{Pineau:2011}  \\ 
\hline
\end{tabular}
\vspace{-10pt}
\end{table}

\vspace{-5pt}
\section{Input methods} \label{im}
\begin{figure}[h]
\centering
\includegraphics[width= 0.9\columnwidth]{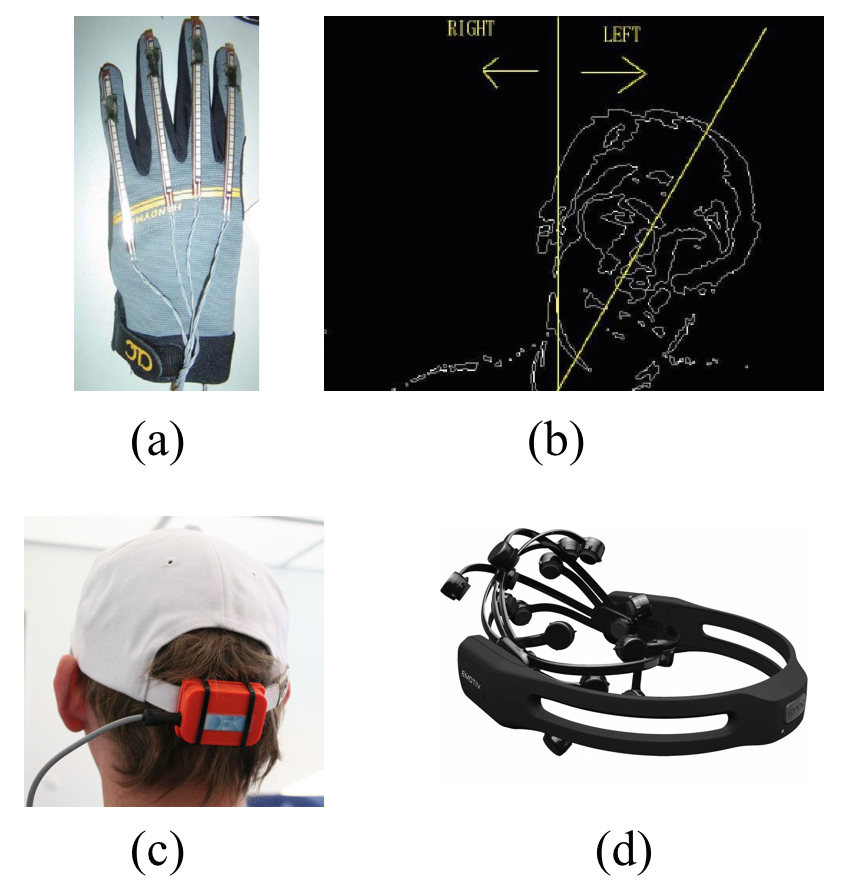}
\caption{Examples of input methods used in SW applications: (a) fingertip control (FTC) \cite{Tyagi2013}, (b) head tilt \cite{Trivedi2013}, (c) accelerometer \cite{Niitsuma2011}, and (d) the Emotiv EPOC headset for Brain computer interface (BCI).}
\label{inputs}
\vspace{-10pt}
\end{figure}

The best choice of input method(s) is not always obvious, but it is usually user specific. Table \ref {tab:im} lists subtopics and example references in the field of SW input method research. Parikh et al. 2007 \cite{Parikh2007} describe the foundations for creating a simple methodology for extracting user profiles, which can be used to adequately select the best command mode for each user. The methodology is based on an interactive wizard with a set of simple tasks for the user, and a method for extracting and analyzing the user's execution of those tasks. Figure \ref {inputs} shows several popular input methods including touch (a), computer vision (b), accelerometer (c), and EEG (d).

Faria et al. 2014 \cite{Faria2014} argue that despite many distinct SW projects around the world, the adaptation of their user interface to the patient is an often neglected research topic. They recommend developing new concepts using multimodal interfaces and wheelchair interfaces adapted to the user's characteristics.

The thought-controlled intelligent machine (TIM \cite{Nguyen2013}) successfully implemented both  head-movement controller (HMC) and  brain-computer interface (BCI). The commercially available electroencephalograph (EEG) Emotiv EPOC headset was evaluated by Carrino et al. in 2012 \cite{Carrino2012}, and at that time could not be used successfully for self-paced applications too sensitive to error.

State-of-the-art BCI can even be used to monitor the user's emotional state \cite{Fattouh2013, Urbano2011}, such that when the user is frustrated the control unit will stop the wheelchair and wait for a new command from the user. In the other case, the control unit will continue executing the previously selected command. 

Deictic control enables the user to move with a series of indications on an interface displaying a view of the environment, bringing about automatic movement of the wheelchair \cite{Leishman2014}. Cloud-based mobile devices \cite{Park2013} have been proposed for direct control of the SW, for remote monitoring, and to help serve as an emergency warning system. Brazil's Hefestos \cite{Tavares2013} has a user interface that communicates with the electric wheelchair firmware via integrated Android app. With it the user can control the wheelchairs' operation from widgets on the touch screen of the user's mobile device.

\begin{figure}[h]
\centering
\includegraphics[width= \columnwidth]{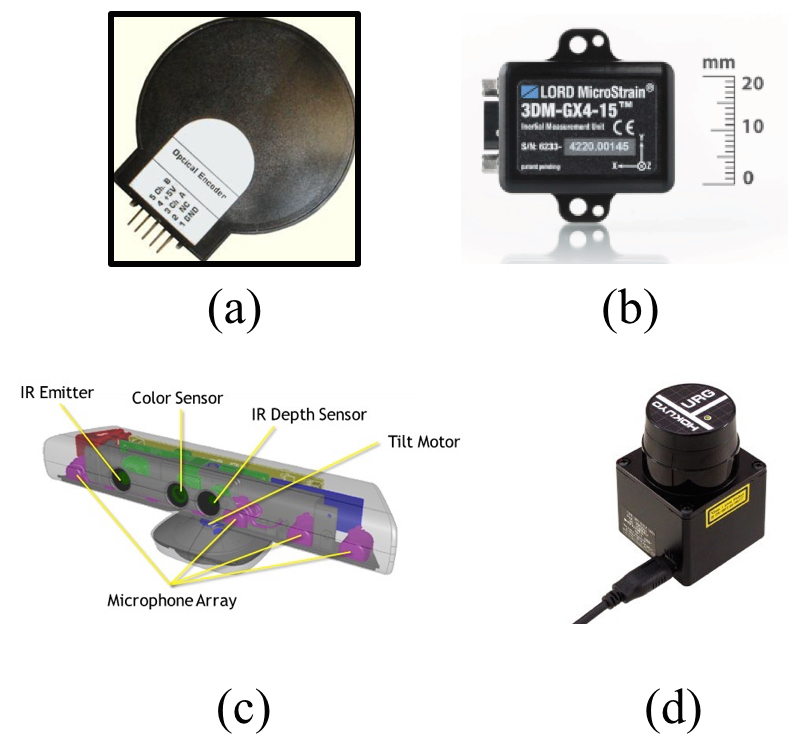}
\caption{Examples of sensors used in SW applications: (a) odometer, (b) inertial measurement unit (IMU) \cite{IMU}, (c) inside Microsoft Kinect \cite{Cockrell2013}, and (d) Hokuyo URG-04LX-UG01 IR laser rangefinder.}
\label{sensors}
\vspace{-10pt}
\end{figure}

Many SWs fuse information from  internal sensors (Odometer, IMU (Fig. \ref{sensors}a,b)) with multiple external sensors to locate obstacles, and provide accurate localization even when one or more of the sensors are operating with large uncertainty. 

A promising method of obstacle detection is a combination of a low-tech, inexpensive optical USB camera and sophisticated machine vision software \cite{Simpson2005}.

Others like Ding et al. \cite{Ding2007} envision a personalized wheelchair navigation system based on a PDA equipped with wireless internet access and GPS that can provide adaptive navigation support to wheelchair users in any geographic environment. Modern smart phones come standard with sensors built in that can be used to supply data about a SW's activities \cite{Milenkovi2013}. 

Using a stereoscopic camera and spherical vision system \cite{Nguyen2012}, 3D scanners like Microsoft's Kinect (Fig. \ref{sensors}c), laser rangefinders (LRFs) \cite{Simpson2005} (Fig. \ref{sensors}d), and most recently the Structure 3D scanner, it has become possible to use point cloud data to detect hazards like holes, stairs, or obstacles \cite{Cockrell2013} (Fig. \ref{ptmap}). These sensors have until recently been relatively expensive, were large and would consume a lot of power.  The Kinect was found to be an effective tool for target tracking, localization, mapping and navigation \cite{Benavidez2011, Kulp2012, Fallon2012}. 

In general, numerous advanced input methods have been developed to serve a wide range of SW users. It would be better if users can control the SW without giving many commands. Therefore, SW research in the future should focus on minimizing the need for users to provide continuous input  commands. SWs should include operating modes that can be trained from users' daily activities to be able to self-operate in most situations.

\vspace{-5pt}
\section{Operating modes}\label{om}
Operating modes range from autonomous to semi-autonomous depending on the abilities of the user and the task at hand. Table \ref {tab:om} lists subtopics and example references in the field of SW operating mode research. Users who lack the ability to plan or execute a path to a destination benefit most from an autonomous system, but only if they spend the majority of their time within the same controlled environment. If the user can effectively plan and execute a path to a destination it may be more beneficial to have a system that is confined to collision avoidance \cite{NavChair, Parikh2007}. Ideally the design should be based around each individual user's abilities and desires, maximizing the amount of control they are given, while offering the user help, as and when it is required \cite{Carlson2008}.

\begin{table}[h]
\caption{Operating modes}
\label{tab:om}
\centering
\begin{tabular}{lc}
\hline
Mode  & References    \\
\hline
Machine learning &  \cite{Tang2012, Touati2012, Tyagi2013} \\
Following  &  \cite{Miyazaki2009, Kobayashi2012, Murakami2014, Suzuki2014,SUGANO2014} \\ 
Localization and mapping  &  \cite{JPLGrids, Wei2012_IRA, Yuki2015_AR,Yang2011, Nguyen2012, Cockrell2013} \\
Navigational assistance  &   \cite{Rofer2009,Viswanathan2011, Poorten2012, Ren2012, Tomari2014, Jain2014} \\ 
\hline
\end{tabular}
\vspace{-10pt}
\end{table}

Among some of the SWs offering semi-autonomous navigation, users may take the additional step of selecting from distinct operating modes depending on the task \cite{Simpson2005}. Operating modes reported to date include:

\vspace{-5pt}
\subsection{Machine learning}
What makes a wheelchair smart is not just a collection of hardware, but specialized computer algorithms that provide the artificial intelligence needed to make split-second decisions about where the wheelchair is heading, and what might be in its way. 

 Some examples of machine learning in SWs include the use of neural networks to detect obstacles \cite{Nguyen2012}, and reproduce pretaught routes \cite{AVNguyen2013}; some use obstacle density histograms to combine information from sonar sensors, and joystick input from the user \cite{NavChair}; while others use rulebased approaches. The LIASD SW has a control architecture that uses the virtual impedance principal \cite{Touati2012}. This approach ensures a smooth implementation of desired tasks by taking into account some real impedance parameter properties such as inertia, viscosity and stiffness. Tyagi et al. 2013 \cite{Tyagi2013} use fuzzy control, which allows the use of inexpensive and imprecise sensors, keeping the overall system cost and complexity low.

Most SWs use reactive control methods, like subsumptive architectures, as the lowest layer of a multilayer architecture \cite{Simpson2005}. This layer interacts directly with the underlying hardware, while upper layers provide deliberative reasoning and control. Tang et al. 2012 \cite{Tang2012} introduce a system that makes extensive use of freely available opensource software. The Player Project has as the base framework: GMapping for mapping, Vector Field Histograms for obstacle avoidance, and Wavefront for path planning.

\vspace{-5pt}
\subsection{Following}
Several SWs \cite{Miyazaki2009, Kobayashi2012, Suzuki2014,SUGANO2014} have been developed that can follow alongside a companion. The SW can track the companion's body position/orientation using laser range sensors based on Kalman filter and data association estimate the guide's footprint. Based on position data of the footprint, a cubic spline generates a target path for the wheelchair. The wheelchair is controlled so that it can move along the guide footprint at a constant gap between the wheelchair and the guide.

In another case, a new methodology was developed \cite{Murakami2014} that allows a wheelchair, which does not have a prior knowledge of the companion's destination, to move with companions collaboratively using a destination estimation model based on observations of human's daily behaviors. 

To serve a large group of people with disabilities at the same time, multi-wheelchair formation controls were developed to enable easy communication with companions \cite{Takano2012}. Controlling the wheelchairs while maintaining formation is done by plotting a path $p$ defined by the model trajectory, and then calculating a path for each wheelchair ($p_{1},p_{2}$) a small distance from $p$ \cite{Iida1991, Akash2014, Li2011}.

\vspace{-5pt}
\subsection{Localization and mapping}\label{ml}

A major challenge is the development of a robust, reliable localization and navigation system in both indoor and outdoor environments. Since the SW needs to safely navigate on roads/sidewalks, it is required that the localization accuracy be within a range of a few centimeters.  The main challenge for outdoor localization is that the global positioning systems (GPS) are not always reliable and robust, especially in tree covered environments. The GPS measurements are integrated with attitude information from the onboard inertial measurement unit (IMU) to enhance the localization accuracy. Moreover, the developed navigation system also fuses the GPS/IMU measurements with the wheel odometry information through an extended Kalman filter (EKF) design \cite{Ohno2004, North2012}. Utilizing the odometry-enhanced fusion, the SW will achieve high-accuracy localization even in GPS-denied environments \cite{La2013TMECH, La2014IROS, LaCASE13_RABIT}.

 \begin{figure}[htb]
\centering
\includegraphics[width= 0.9\columnwidth]{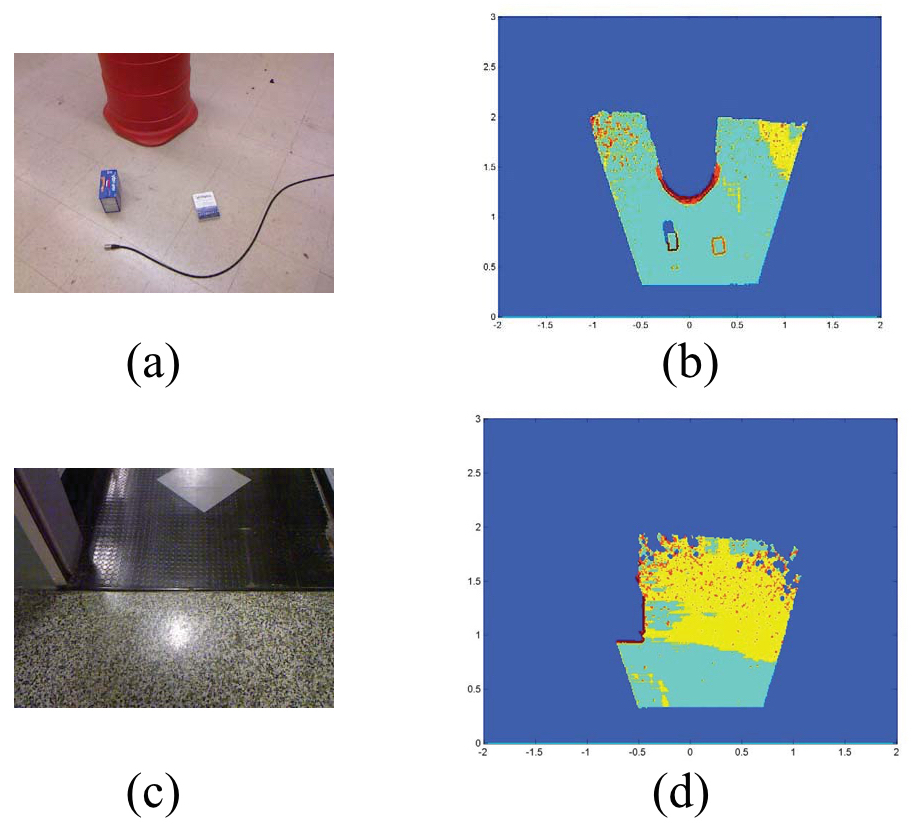}
\caption{Determining navigability of terrain using point cloud data from a Microsoft Kinect \cite{Cockrell2013}: (a) RGB image floor type 1 (F1) with an obstacle and a few bumps. (b) Point cloud map of F1 indicating obstacles with red, flat safe terrain in light blue, and bumps in shades of orange and yellow. (c) RGB image floor type 2 (F2) with a doorpost and a terrain change from smooth to bumpy. (d) Point cloud map of F2.}
\label{ptmap}
\vspace{-5pt}
\end{figure}
 \begin{figure}[htb]
\centering
\includegraphics[width= \columnwidth]{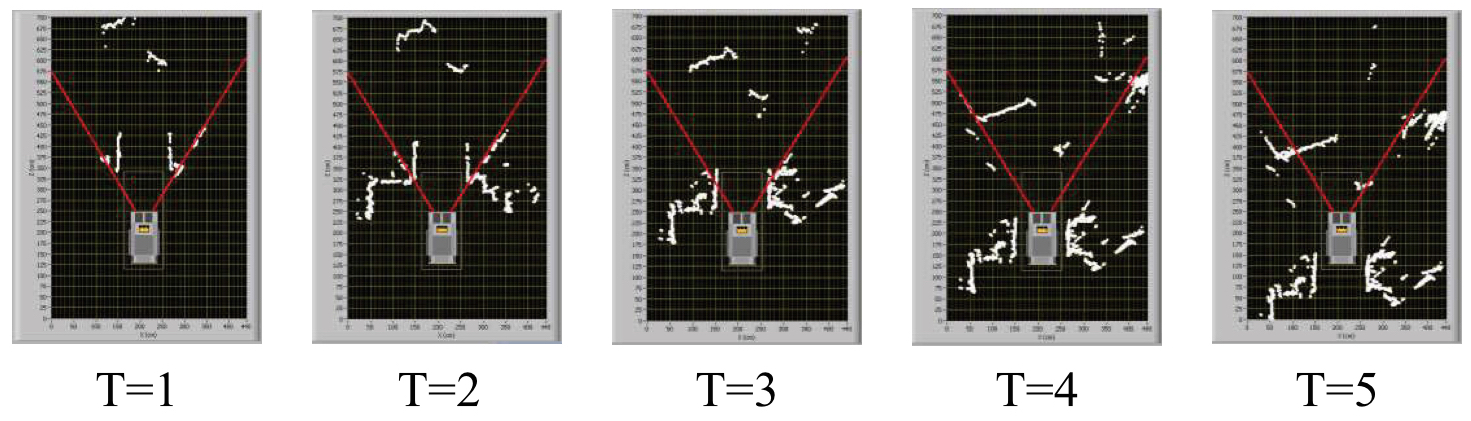}
\caption{2D birds eye view of the environment derived from stereoscopic observations with the Bumblebee XB3 \cite{bumblebee}. Object points once plotted are remembered even when out of range of vision \cite{Nguyen2012}.}
\label{2dmap}
\vspace{-10pt}
\end{figure}
 \begin{figure}[!t]
\centering
\includegraphics[width=0.9\columnwidth]{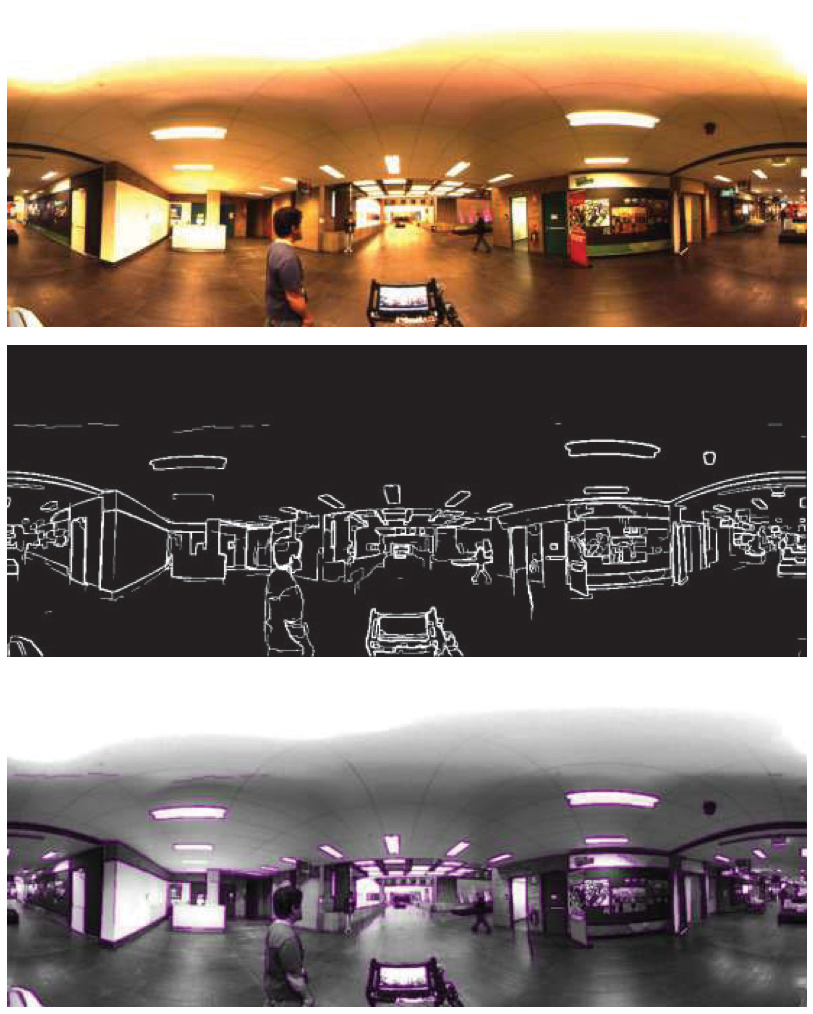}
\caption{Top: panoramic image captured using a BGR Ladybug2 \cite{ladybug} spherical vision camera; Middle: after applying the improved Prewitt edge detection method \cite{Yang2011}; Bottom: edge overlay on grayscale of the original panoramic image \cite{Nguyen2012}.}
\label{edges}
\vspace{-10pt}
\end{figure} 

Fig. \ref{ptmap} shows an example of obstacle detection using Microsoft's Kinect \cite{Cockrell2013}. An RGB image of a floor with an obstacle and a few bumps is shown side-by-side with a point cloud map of the same floor. Obstacles are red, flat safe terrain is light blue, and bumps are shades of orange and yellow. 

3D point cloud data can also be displayed as a 2D birds eye view of the SW's environment \cite{Nguyen2012} (Fig. \ref{2dmap}). From left to right the time sequence (T1-T5) illustrates the SW advancing into a previously uncharted room, continuously building a map, and keeping a record of obstacles for later reference when they are out of range of vision. Between frames 1 and 2 the pilot turns side to side to collect data.

Using spherical vision cameras like the BGR Ladybug2 \cite{ladybug} and applying methods like the improved Prewitt edge detection \cite{Yang2011}, SWs can detect obstacles in most indoor environments \cite{Nguyen2012} (Fig. \ref{edges}). 

\vspace{-5pt}
\subsection{Navigational assistance}

An elegant application of semi autonomous obstacle avoidance was accomplished by Rofer et al. 2009 \cite{Rofer2009} for a rear-wheel drive SW, where the front active castor wheels are always rotated by motors to match the orientation of the current driving direction.

Viswanathan et al 2011. \cite{Viswanathan2011} run multiple processes in a distributed fashion on a Nimble Rocket $^{TM}$ that serve people with visual impairments particularly well. 

\begin{itemize}
\item{Collision Detector: detects frontal collisions and stops the wheelchair if an object is detected within a distance of approximately 1 meter, preventing motion in the direction of the obstacle through the controller.}
\item{Path Planner: given a global map of the environment and an initial position estimate, visual odometry produce the optimal route to the specified goal location. The trajectory is analyzed to determine deviations from the optimal route as well as upcoming turns.}
\item{Prompter: uses a Partially Observable Markov Decision Process (POMDP) to determine the optimal prompting strategy. Specifically, this module estimates the users' ability level to navigate to the goal independently based on past errors, and responsiveness to prompts in order to select appropriate audio prompts to assist the users in navigation.}
\end{itemize}

Jain et al. 2014 \cite{Jain2014} present an algorithm for the automated detection of safe docking locations at rectangular and circular docking structures (tables, desks) with proper alignment information using 3D point cloud data. Based on geometric information, computed from depth data, the method is invariant to scene or lighting changes. It does not make use of fiducial landmarks, or simplify the problem by using customized docking structures \cite{Ren2012}. The safe docking locations can then be provided as goals to an autonomous path planner. 

The same docking principals also apply to the joining of a SW with a transport vehicle like a van or bus. Gao et al. 2007 \cite{Gao2007} devised a LIDAR based Automated Transport and Retrieval System (ATRS) as an alternative to van conversions for automobile drivers.

Poorten et al. 2012 introduces a set of novel haptic guidance algorithms intended to provide intuitive and reliable assistance for PW navigation through narrow or crowded spaces. Tomari et al. 2014 \cite{Tomari2014} propose a method of navigation in crowded indoor environments based on the observation of people's head information obtained from color and range images. Initially head regions in the color image are tracked and their orientations are estimated, then the head orientation data are integrated with the head position data in the range image to determine the wheelchair motion so that it can smoothly move among humans.

In each operating mode, a huge number of achievements have been made to improve the SW's functionality. Due to current sensor capabilities limiting autonomous navigation to indoor environments, combining all of the above advancements can result in autonomous navigation through all environments. Thus, autonomous navigation would be one of the main focuses of  future SW research. Furthermore, some SW users still significantly depend on caregivers, especially people with quadriplegia. While robotic technology will not completely replace human caregivers, it can help to extend the time that an individual can spend in independent living situations \cite{Padir2014}.

\vspace{-5pt}
\section{Human Factors in Smart Wheelchair}\label{hf}
As the population of industrialized countries ages, the number of adults living with mobility impairment is increasing, with an estimated 2.8 million Americans residing outside of institutions utilize wheelchairs as an assistive technology device \cite{RESNA2011}.  This population includes, but is not limited to, individuals with low vision, visual field neglect, spasticity, tremors, or cognitive deficits. Challenges to safe, independent wheelchair use can result from various overlapping physical, perceptual, or cognitive symptoms of diagnoses such as spinal cord injury, cerebrovascular accident, multiple sclerosis, amyotrophic lateral sclerosis, and cerebral palsy.  Persons with different symptom combinations can benefit from different types of assistance from a SW and different wheelchair form factors \cite{Gerling2014, Carrington2014, Narayanan2014}. Manual and powered wheelchairs are standard approaches to address such impairments, but these technologies are unsuitable for many users \cite{Mitchell2014}. While the needs of some individuals with disabilities can be satisfied with PWs, many people with disabilities (about 40\% of the disabled community \cite{Fehr:2000}) find operating a standard PW difficult or impossible (Fig. \ref{fall}).

In the past SWs have simply been considered mobile robots that are transformed ``by adding a seat". The main difference between a SW and a robot with a seat is that a wheelchair is an extension of the human being sitting in it and he or she must feel comfortable using it. Thus, human factors are a crucial component in the design and development of a SW \cite{Carlson2013_RAM, Gunachandra2014, Jang2014_IROS, Widyotriatmo2015_ASCC}. Table \ref {tab:hf} lists subtopics and example references in the field of SW human factor research.

\begin{table}[h]
\caption{Human factors}
\label{tab:hf}
\centering
\begin{tabular}{lc}
\hline
Factor   & References    \\
\hline
Human Robot Interaction &  \cite{Hancock2011, Riek2014} \\
 Interface  &  \cite{Alrajhi2014} \\
  Learning  &  \cite{CanWheel, Gironimo2013} \\
   Operations  &  \cite{Adhikari2014} \\
    Physiology  &  \cite{Jipp2012, KANKI2014, Giesbrecht2014} \\
     Platform  &  \cite{Shiomi2014} \\
      Social issues  &  \cite{Jensen2014}\\
\hline
\end{tabular}
\vspace{-18pt}
\end{table}

\vspace{-5pt}
\subsection{Human Robot Interaction}
A recent meta-analysis of the available literature on trust and human-robot interaction (HRI) found that robot performance and attributes were the largest contributors to the development of trust in HRI. Environmental factors played only a moderate role \cite{Hancock2011}.

In a discussion on the unique technical, ethical, and legal challenges facing HRI practitioners when designing robots for human social environments, Riek et al. 2014 argue that the affordance of all rights and protections ordinarily assumed in human-human interactions apply to human-robot interaction, and propose a code of ethics for the profession \cite{Riek2014}.

 Robots intended to provide physical assistance to people with disabilities present a unique set of ethical challenges to HRI practitioners. Robots may be used for daily living tasks, such as bathing, manipulation, mobility and other activities to support independent living and aging-in-place. The design of physically assistive robots must, therefore, take into consideration the privacy rights of clients, as with, perhaps, the deactivation of video monitors during intimate procedures. 
  
  \vspace{-5pt}
  \subsection{Learning}
  A team of 18 clinical researchers at the University of British Columbia have developed a program of research that will use a mixed-methods approach, including lab-based, qualitative and quantitative methodologies, spanning 5 key research projects. Their plan is to assess the effectiveness of their new, collaboratively-controlled wheelchair, in combination with the Wheelchair Skills Program \cite{CanWheel}.
  
 Gironimo et al. 2013 at the University of Naples use virtual reality based experiments in which they collect quantitative measures of users operating a wheelchair-mounted robot manipulator, followed by subjective user evaluations intended to maximize the effectiveness, the efficiency, and the satisfaction of users \cite{Gironimo2013}. 
 
      Even if stable and safe wheelchair robots are realized, integrating them in the world is difficult without acceptance by the actual users: elderly people. Experimental results with elderly participants at a pseudo resident home show that the seniors prefer wheelchair robots programed to make appropriate utterances to reduce their anxieties. Seniors also preferred an adaptable wheeling speed, since speed preference varies for each individual \cite{Shiomi2014}.
 
 \vspace{-5pt}     
   \subsection{Physiology}
\begin{figure*}[t!]
\centering
\includegraphics[width=\textwidth]{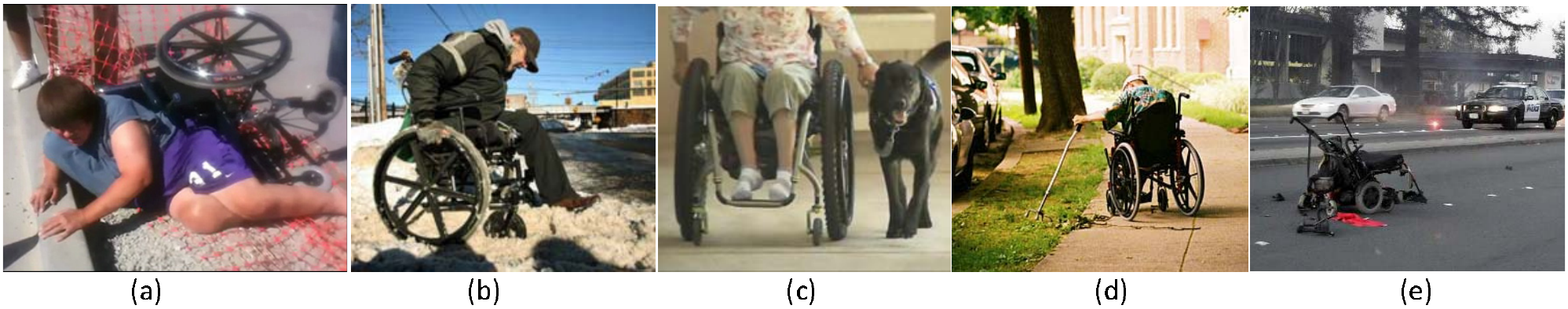}
\caption{Navigation problems for wheelchair users: (a,b) Accidents with manual navigation for visually impaired wheelchair users; (c) Service-dog for navigation support; (d) Navigation with cane holder; (e) Accident happened due to poor navigation.}
\label{fall}
\vspace{-10pt}
\end{figure*}

   PW users are involved in some accidents once in a while. Most common are falls off of the wheelchair due to inclined surfaces (Fig. \ref{fall}-a,b), so researchers at the Kyoto Institute of Technology measure the physiological indices when the wheelchair is inclined, in order to evaluate the physical strains on the human body \cite{KANKI2014}. Their results show that even though the contact area does not get higher as the incline angle gets higher, the distribution of high pressure area gets lower as the angle gets higher, therefore people feel uncomfortable as the inclination gets higher.
  
Older adults typically receive little or no mobility training when they receive a wheelchair, and are therefore more likely to experience an accident. In order to accommodate this underserved population Giesbrecht et al. 2014 \cite{Giesbrecht2014} developed a therapist-monitored wheelchair skills home training program delivered via a computer tablet. To optimize efficacy and adherence, principles of self-efficacy and adult learning theory were foundational in the program design. 

A study with 23 participants by Meike Jipp of the German Aerospace Center in 2012 found that participants with lower fine motor abilities had more collisions and required more time for reaching goals. Adapting the wheelchair's level of automation to these fine motor abilities can improve indoor safety and efficiency \cite{Jipp2012}.
 
 \vspace{-5pt}
   \subsection{Social issues}
A 2014 study of data in the form of 55 letters from 33 Danish citizens applying for assistive technology established the following as core aspects of user perspectives on assistive devices: quality of life, realization of personal goals, keeping up roles, positive self image, personal dignity, independence, activities and participation, self-care and security, lack of access in surroundings, professional domination of assistive device area, cultural stigma of negative symbolic value, and cumbersome standardized devices \cite{Jensen2014}.

  The program identifies several factors that will impede participation in public activities by a person with a mobility impairment. 
  
  \begin{itemize} 
  \item{Individual: compromised body function and structure related to pain, strength, and endurance.} 
  \item{Built environment: inaccessible locations, and challenging terrain.} 
  \item{Social environment: social attitudes and level of personal assistance.} 
   \item{Assistive technology devices including wheelchairs that are low-quality and inappropriate or do not fit the user.}
    \item{Contextual factors: age and confidence (self-efficacy) with wheelchair use.}
   \end{itemize}      

\vspace{-5pt}
\subsection{Commercialization}\label{pc}
There are three reasons to purchase a SW, each requiring a different behavior: 

\begin{enumerate}
\item{Mobility aid, helps the user reach a destination as quickly and comfortably as possible.}
\item{Training tool, helps the user develop specific skills. The amount of user input required is a function of the training activity.} 
\item{Evaluation instrument, records activity without intervention.}
\end{enumerate}

\begin{table}[h]
\caption{Commercially available 'smart' wheelchairs. Unfortunately, companies call their product smart even though there are no sensors, computers, or novel software.}
\label{tab:availSW}
\centering
\begin{tabular}{llcc}
\hline
Name  & Novelty  & Smart (Y/N) & Reference    \\
\hline
 EZ Lite Cruiser & PW that folds &   N &  \cite{EZlitecruiser} \\
 KD smartchair & PW that folds &   N &  \cite{KDsmartchair} \\
 SmartDrive MX2 & Converts W to PW &   N &  \cite{smartdrive} \\
 Smile Rehab's SW & Follows a line & Y & \cite{SRsmartchair} \\
  & Avoids obstacles & & \\
\hline
\end{tabular}
\end{table}

Smile Rehab's SW is the only SW available to the consumer (see Table \ref{tab:availSW}), because the technology is still expensive, and sensor capabilities limit autonomous navigation to indoor environments, that have to be modified by putting down tape. Another downside is that the consumer has to use Smile Rehab's PW platform. 

Henderson et al. 2014 \cite{Henderson2014} are providing researchers with an embedded system that is fully compatible, and communicates seamlessly with current manufacturer's wheelchair systems. Their system is not commercially available, but researchers can mount various inertial and environmental sensors, and run guidance and navigation algorithms which can modify the human desired joystick trajectory, to help users avoid obstacles, and move from room to room. SWs should be customized to each consumer, making it really difficult to expand a business or lower the cost of a SW (by mass production).

Toyota Motor East Japan Inc. \cite{Patrafour}, a manufacturing subsidiary of Toyota Motor Corporation, has gotten into the personal electric vehicle (PEV) business (Fig. \ref{forms}e), which demonstrates that there is a market for technology very similar the ideal SW, a PEV modified for use indoors by an individual with a disability.

In general, apart from current technological issues, non-technical barriers also hamper commercialization. Prescribers and insurance providers want to avoid liability, and want proof that the SW technology works. A SW may be tested a thousand times without a user, under laboratory conditions, but it is more difficult to find a real human subject to participate in the trials. With very few trial participants, data has been difficult to collect, and insurers reluctant to provide reimbursement. Meanwhile, policymakers are concerned about the increasing demand for unnecessarily elaborate wheelchairs \cite{Greer2012}, and since SW studies are considered human trials, they are subject to strict government regulations \cite{hhs}. 

\vspace{-5pt}
\section{Future: Co-robot}\label{Future}
A study conducted by Worcester Polytechnic Institute (WPI) \cite{Padir2014} showed that users who are accustomed to their wheelchairs have little to no tolerance for the failures of ``new features,'' which means full autonomy plays an important role in users' acceptability. 
Overall, people are open to using a robot that assists them in performing a task in an environment with other humans. These "Co-robots" provide their user with independent mobility.  
SWs have to be safe and their operations must be reliable under any conceivable circumstance. This is especially true for SWs, where the user has a disability and may be quite frail. To emphasize these trends, we propose and discuss the following  future  directions which would be the priorities for SW research.

\vspace{-5pt}
\subsection{Autonomous Navigation}
Current PW users still have many difficulties with daily maneuvering tasks (Fig. \ref{fall}), and would benefit from an automated navigation system. Future SW users will be able to select from several autonomous or semi-autonomous operating modes, using the input methods they are most comfortable with. They will direct their SW to navigate along preprogrammed routes with little physical or cognitive effort indoors and out, passing through doors and up-and-down elevators. The SW will communicate with the user when it is appropriate in order to reduce anxiety, and build an individualized profile. These profiles will track such variables as preferred input method, wheeling speed, turning speed, and amount of verbal feedback, just to name a few. 

\begin{figure}[h!]
 \centering
 \includegraphics[width = \columnwidth]{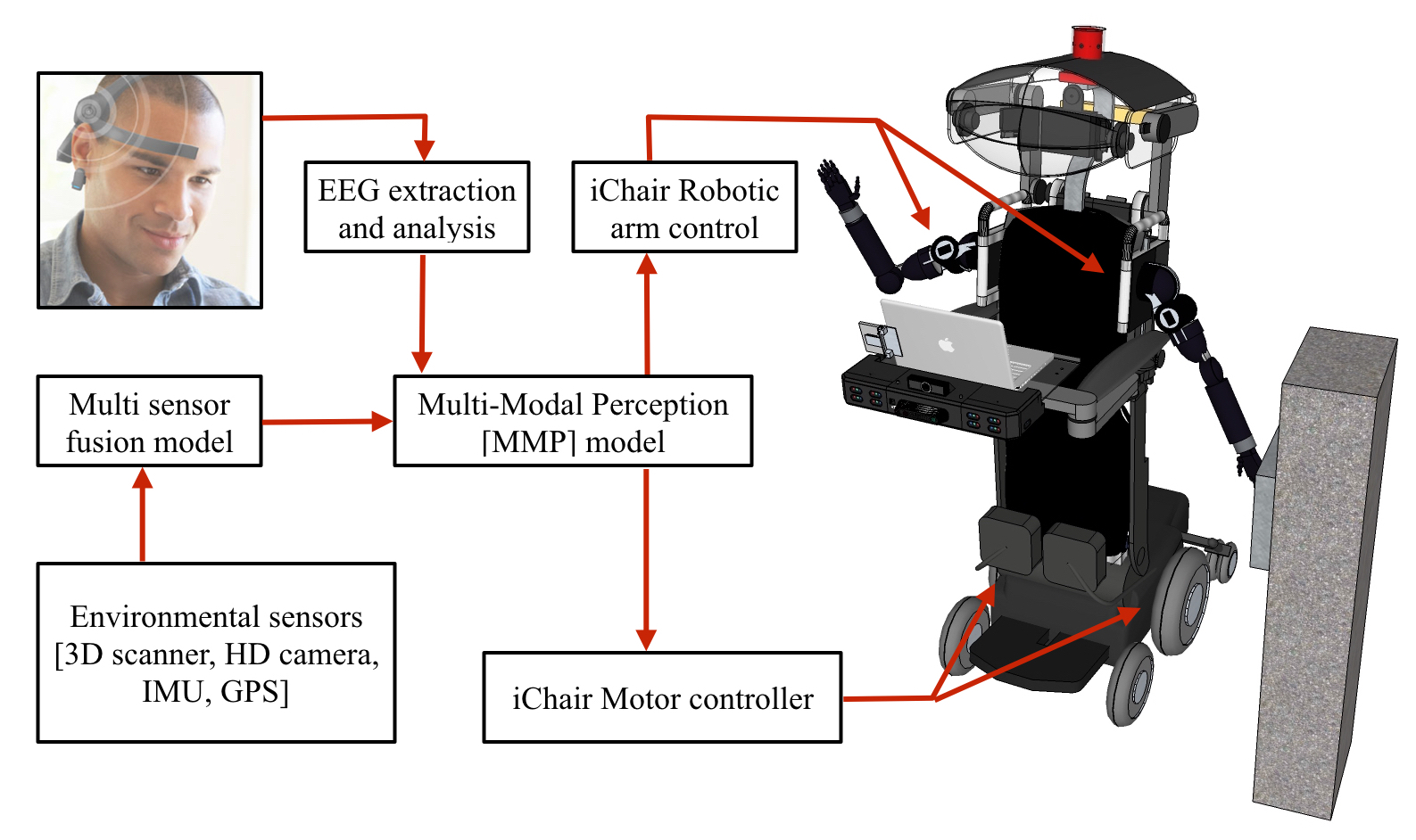}
  \caption{Example diagram of the human-iChair interaction model.}
  \label{human-chair}
\vspace{-10pt}
\end{figure}

One of the downsides of the earlier work in SWs and autonomous driving \cite{Simpson:2004, Bourhis2001, Simpson:2008, Pineau:2011, Kobayashi2012, Pretz2013, Murakami2014, Suzuki2014, Rathore2014, Rocha2014, Sinyukov2014} has been that the environment had to be permanently modified in order for the wheelchair to navigate independently. Future research plans should focus on developing novel navigation algorithms to allow SWs to autonomously and safely navigate in complex environments. These are not trivial tasks since they require online calibration \cite{ La2011ICRA, La2014TASE} and sensor fusion of multiple sensor sources including laser scanners, camera, IMU, encoder and GPS, etc. Fortunately, the navigation algorithms developed for SWs could inherit the advanced performances of previous works on autonomous navigation for the robots \cite{La2013TMECH,  La2014IROS} and or autonomous cars.

\vspace{-5pt}
\subsection{Human-Smart Wheelchair Interaction Model }
The human-SW interaction model plays a vital role in the formation of a SW. The reinforcement learning technique \cite{Sutton1998, Reis2014} could be utilized to build an efficient interaction model between user and SW. The developed model may take into account sensor feedbacks, such as from Emotive sensors \cite{Emotiv}  and various environment sensing sensors including Oculus virtual reality sensors, laser scanners, cameras and global positioning system (GPS).
 The SW uses these control signals to navigate and operate its robotic arms (see Fig. \ref{human-chair}). 
 Human user thinking will be captured and analyzed through an advanced signal processing model \cite{Sinyukov2014}.  
The output signals will control the operations of the SW including wheel motors as well as robotic arms. A shared control scheme could be developed to allow effective collaboration between user and SW, as well as to adapt SW's assistance to the variations in user performance and  environmental changes.

 A real-time Multi-Modal Perception (MMP) model should be used to deal with uncertainty in sensing, which is one of the most enduring challenges. The perceptual challenges are particularly complex, because of the need to perceive, understand, and react to human activity in real-time. The range of sensor inputs for human interaction is far larger than for most other robotic domains in use today. MMP inputs include human thought, speech and vision, etc., which  are major open challenges for real-time data processing. Computer vision methods  processing human-oriented data such as facial expression \cite{Peters2013} 
 and gestures recognition \cite{Drumwright2004} will need to be developed to handle a vast range of possible inputs and situations. Additionally, language understanding and dialog systems between human users and SWs have to be  investigated \cite{Narayanan2007, Obaidat2014}. 
 MMP can obtain an understanding of the connection between visual and linguistic data \cite{Rizzolatti1998, Akash2014} and combine them to achieve improved sensing and expression with a SW.

\begin{figure}[htb]
\centering
\includegraphics[width= \columnwidth]{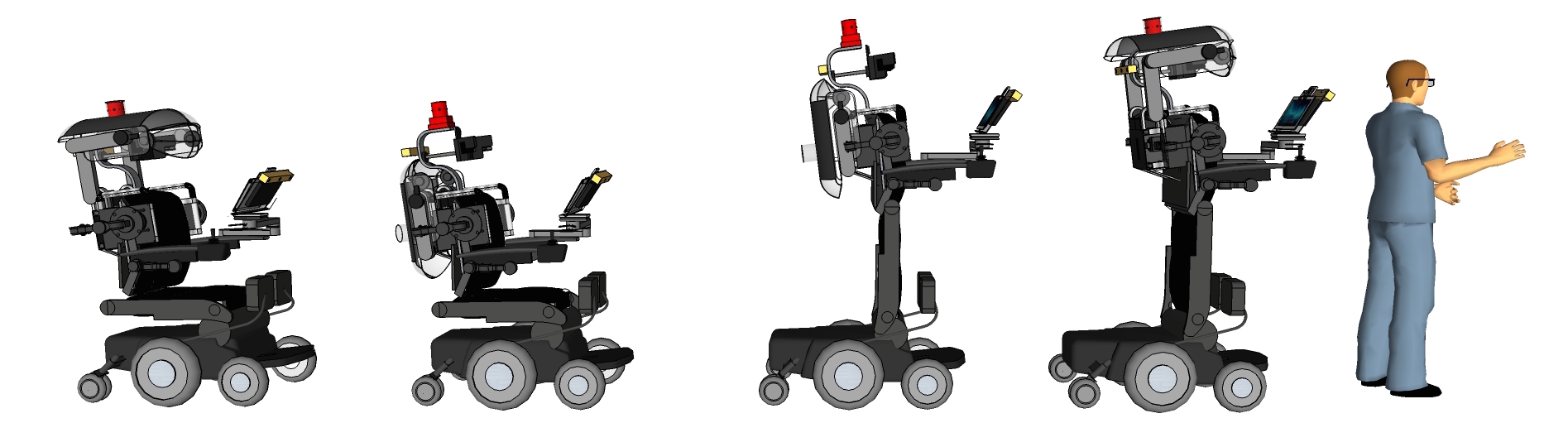}
\centering
\caption{From left to right: four iChair Australia Editions (AE)s, two sitting, one with roof up, one with roof down, and two standing, one with roof down and the other up, all in I-formation behind a human guide. The iChair AE includes the Retractable Roof with Heads-up Display (RR$\&$ HD),  Robotic Arm Assistant (RAA),  Ladybug2 and 2 Bumblebee sensors, mounted on a Levo C3 power wheelchair with standing ability. Shown with joystick, HCI and tablet running voice control, head tracking, and an Android app with a menu of options such as: drive options, maps, sensors, RR, HD, RAA.}
\label{human chair}
\vspace{-15pt}
\end{figure}

\subsection{Smart Wheelchair with Smart Home}
SWs will be integrated into the smart home, providing a seamless user experience and control over all household appliances. When the users are outdoor a retractable roof will provide shelter from the elements, and additional safety at night while driving in traffic. Optical stereoscopic and spherical vision imagery will be combined with infrared laser data to produce a virtual point cloud matrix of the user's surroundings. Objects in the matrix are identified using machine vision, visual tracking, and gesture recognition algorithms. Localization information from the IMU and GPS, onboard data collected, and bluetooth beacon data flooding public spaces, will all help guide the SW to a particular destination. Fig. \ref{human chair} is a 3D rendering of a standing SW with a retractable roof and robotic arms that will allow users with disabilities of all types to independently perform many more daily living activities. 

We believe that given liability concerns, SWs in the future should be treated like PEVs that are registered and insured. Users should be certified, or licensed, for operation after passing a standardized wheelchair skills test \cite{Pineau:2011, CanWheel}. 

\vspace{-5pt}
\section{Conclusion}\label{Con}
SWs represent a paradigm shift unlike the one that occurred when power wheelchairs became available on a mass scale, which was a technological advancement from the manual wheelchair. In the last decade sensors and computers have gotten faster, cheaper, and smaller, while computer vision software has become more sophisticated and readily available than ever before. The research community has developed many prototypes, the best parts of which should be fused into a modular upgradable system that can be marketed to the millions of people who need PWs. Electric vehicle manufacturers are in a prime position to produce a stand-alone SW, and capitalize on the millions of aging Babyboomers worldwide, who will want to have a SW. 

The best SWs will be able to accommodate people with all disability types by utilizing a multi-modal interface that combines computer vision, touch, voice, and brain control. SWs will be able to build 3D maps using mobile scanners, and navigate autonomously by streaming and analyzing sensory data in real-time through cloud computing applications. These solutions show much promise to significantly improve the quality of life for people who use PWs, but only if they are trusted by the people they were designed to serve. This means that human factors have be taken into account, so SWs should be customizable to the individual user's preferences, provide verbal feedback when appropriate, and be mountable on any PW. The stage is set for humans and robots to interact in public spaces in order to give people with disabilities the best quality of life possible, and the opportunity to maximize their human potential.
\vspace{-15pt}
\section*{Acknowledgment}
The authors would like to thank Nhan Pham for his help to proofread the paper and re-organize the references and search keywords.
\bibliographystyle{unsrt}
\bibliography{References6}

\begin{thebibliography}{100}

\bibitem{Mazumder2014}
O.~Mazumder, A.S. Kundu, R.~Chattaraj, and S.~Bhaumik.
\newblock Holonomic wheelchair control using {EMG} signal and joystick
  interface.
\newblock In {\em Recent Adv. in Eng. and Comput. Sci.}, pages 1--6,
  Chandigarh, India, Mar. 2014.

\bibitem{pasteau2014}
F.~Pasteau, A.~Krupa, and M.~Babel.
\newblock {Vision-based assistance for wheelchair navigation along corridors}.
\newblock In {\em {IEEE Int. Conf. Robot. and Auto.}}, pages 4430--4435,
  Hong-Kong, Hong Kong SAR China, Jun. 2014.

\bibitem{Desmond2013_TePRA}
R.~Desmond, M.~Dickerman, J.~Fleming, D.~Sinyukov, J.~Schaufeld, and T.~Padir.
\newblock Develop. of modular sensors for semi-autonomous wheelchairs.
\newblock In {\em IEEE Int. Conf. Technol. for Practical Robot Applicat.},
  pages 1--6, Woburn, MA, Apr. 2013.

\bibitem{Sinyukov2014}
D.~Sinyukov, R.~Desmond, M.~Dickerman, J.~Fleming, J.~Schaufeld, and T.~Padir.
\newblock Multi-modal control framework for a semi-autonomous wheelchair using
  modular sensor designs.
\newblock {\em Intell. Service Robot.}, 7(3):145--155, Jul. 2014.

\bibitem{Millan2014_BIC}
J.~d.~R.~Millan.
\newblock {BMI: Lessons from tests with impaired users}.
\newblock In {\em Int. Winter Workshop Brain-Comput. Interface}, pages 1--1,
  Jeongsun-kun, Feb. 2014.

\bibitem{Rathore2014}
D.K. Rathore, P.~Srivastava, S.~Pandey, and S.~Jaiswal.
\newblock A novel multipurpose smart wheelchair.
\newblock In {\em IEEE Students' Conf. Elect., Electron. and Comput. Sci.},
  pages 1--4, Bhopal, Mar. 2014.

\bibitem{Yayan2014}
U.~Yayan, B.~Akar, F.~Inan, and A.~Yazici.
\newblock Develop. of indoor navigation software for intelligent wheelchair.
\newblock In {\em IEEE Int. Symp. Innovations in Intell. Syst. and Applicat.
  Proc.}, pages 325--329, Alberobello, Jun. 2014.

\bibitem{Leishman2014}
F.~Leishman, V.~Monfort, O.~Horn, and G.~Bourhis.
\newblock Driving assistance by deictic control for a smart wheelchair: {T}he
  assessment issue.
\newblock {\em IEEE Trans. Human-Mach. Syst.}, 44(1):66--77, Feb. 2014.

\bibitem{Jain2014_IROS}
S.~Jain and B.~Argall.
\newblock Automated perception of safe docking locations with alignment
  information for assistive wheelchairs.
\newblock In {\em IEEE/RSJ Int. Conf. Intell. Robots and Syst.}, pages
  4997--5002, Chicago, IL, Sept. 2014.

\bibitem{Simpson:2004}
R.~Simpson.
\newblock {Smart Wheelchair Component System}.
\newblock {\em J. Rehabil. Research and Develop.}, 41(3B):429--442, 2004.

\bibitem{Pineau:2011}
J.~Pineau, R.~West, A.~Atrash, J.~Villemure, and F.~Routhier.
\newblock {On the Feasibility of Using a Standardized Test for Evaluating a
  Speech-Controlled Smart Wheelchair}.
\newblock {\em Int. J. Intell. Control and Syst.}, 16(2):124--131, 2011.

\bibitem{Chao2008_JFR}
C.~Gao, T.~Miller, J.~R. Spletzer, I.~Hoffman, and T.~Panzarella.
\newblock Auton. docking of a smart wheelchair for the automated transport and
  retrieval system (atrs).
\newblock {\em J. Field Robot.}, 25(4-5):203--222, May 2008.

\bibitem{Andrew2010_FSR}
C.~Gao, M.~Sands, and J.R. Spletzer.
\newblock {\em Field and Service Robot.: Results of the 7th Int. Conf.},
  chapter Towards Auton. Wheelchair Syst. in Urban Environments, pages 13--23.
\newblock Springer Berlin Heidelberg, Berlin, Heidelberg, 2010.

\bibitem{Li2013}
R.~Li, L.~Wei, D.~Gu, H.~Hu, and K.D. McDonald-Maier.
\newblock Multi-layered map based navigation and interaction for an intell.
  wheelchair.
\newblock In {\em Int. Conf. Robot. and Biomimetics}, pages 115--120, Shenzhen,
  China, Dec. 2013.

\bibitem{Simpson2005}
R.~Simpson.
\newblock Smart wheelchairs: A literature review.
\newblock {\em J. Rehabil. Research and Develop.}, 42(4):423--436, Aug. 2005.

\bibitem{Ding2005_CS}
D.~Ding and R.A. Cooper.
\newblock Electric powered wheelchairs.
\newblock {\em IEEE Control Syst.}, 25(2):22--34, Apr. 2005.

\bibitem{iBot}
Deka-Research.
\newblock Stair climbing wheelchair {iBot}.
\newblock {\em http://www.dekaresearch.com/ibot.shtml}, Feb. 2015.

\bibitem{abilityinmotion}
Ability in~Motion.
\newblock Standing wheelchair.
\newblock {\em http://abilityinmotion .com.au/products/standing-wheelchairs/},
  Feb. 2015.

\bibitem{tankchair}
TC-Mobility.
\newblock Power chair with tank chassis.
\newblock {\em http://www.tankchair.com}, Feb. 2015.

\bibitem{observer}
Observer-Mobility.
\newblock All-terrain wheelchair.
\newblock {\em http://www.observer-mobility.com}, Feb. 2015.

\bibitem{Patrafour}
Toyota Motor~East Japan.
\newblock Patrafour.
\newblock {\em http://www.toyota-ej.co.jp/english/products/patrafour.html},
  Jul. 2015.

\bibitem{Globe2012}
M.~Yukselir, K.~Scarrow, P.~Wilson, and T.~Cowan.
\newblock The brains behind the electric wheelchair, one of {Canada's} 'great
  artifacts'.
\newblock {\em The Globe and Mail}, Aug. 27, 2012.

\bibitem{EJ1956}
K.~Fritsch.
\newblock The neoliberal circulation of affects: Happiness, accessibility and
  the capacitation of disability as wheelchair.
\newblock {\em Health, Culture and Soc.}, 5(1), 2013.

\bibitem{Smartchair}
KD-Healthcare-Company.
\newblock Affordable, foldable motorized wheelchair.
\newblock {\em http://kdsmartchair.com}, Feb. 2015.

\bibitem{LeamanNews}
M.~Chandler.
\newblock America's next great inventions.
\newblock {\em Mercury News, Available at:
  http://www.mercurynews.com/c-i5485125}, Mar. 2007.

\bibitem{iChair}
J.~Leaman and H.~M. La.
\newblock {iChair}: Intell. powerchair for severely disabled people.
\newblock In {\em ISSAT Int. Conf. Modeling of Complex Syst. and Environments},
  Da Nang, Vietnam, Jun. 2015.

\bibitem{Leaman_MFI16}
J.~Leaman, H.~M. La, and L.~Nguyen.
\newblock Development of a smart wheelchair for people with disabilities.
\newblock In {\em IEEE Int. Conf. on Multisensor Fusion and Integration for
  Intell. Sys. (MFI)}, pages 279--284, 2016.

\bibitem{Headmouse}
NaturalPoint.
\newblock Smartnav head tracking mouse.
\newblock {\em http://www.naturalpoint.com/smartnav/}, Feb. 2015.

\bibitem{MountnMover}
BlueSkyDesigns.
\newblock Mount-n-mover.
\newblock {\em http://www.mountnmover.com/ product-options/dual-arm/}, Feb.
  2015.

\bibitem{VAHM}
G.~Bourhis, K.~Moumen, P.~Pino, S.~Rohmer, and A.~Pruski.
\newblock Assisted navigation for a powered wheelchair. syst. eng. in the
  service of humans:.
\newblock In {\em Proc. IEEE Int. Conf. Syst., Man and Cybern.}, pages
  553--558, Le Touquet, France, Oct. 1993.

\bibitem{MrED}
J.~Connell and P.~Viola.
\newblock Cooperative control of a semi-autonomous mobile robot.
\newblock In {\em Proc. IEEE Int. Conf. Robot. and Auto.}, pages 1118--1121,
  Cincinnati, OH., May 1990.

\bibitem{Simpson:2008}
R.~Simpson.
\newblock {How many people would benefit from a smart chair?}
\newblock {\em J. Rehabil. Research and Develop.}, 45(1):53--72, 2008.

\bibitem{Navysmartchair}
S.~P. Parikh, V.~Grassi Jr, V.~Kumar, and J.~Okamoto Jr.
\newblock Usability study of a control framework for an intell. wheelchair.
\newblock In {\em Proc. IEEE Int. Conf. Robot. and Auto.}, pages 4745--4750,
  Barcelona, Spain, Apr. 2005.

\bibitem{Gao2007}
C.~Gao, I.~Hoffman, T.~Miller, T.~Panzarella, and J.~Spletzer.
\newblock Performance characterization of lidar based localization for docking
  a smart wheelchair system.
\newblock In {\em IEEE/RSJ Int. Conf. Intell. Robots and Syst.}, San Diego, CA,
  Nov. 2007.

\bibitem{Ding2007}
D.~Ding, B.~Parmanto, H.~A. Karimi, D.~Roongpiboonsopit, G.~Pramana,
  T.~Conahan, and P.~Kasemsuppakorn.
\newblock Design considerations for a personalized wheelchair navigation
  system.
\newblock In {\em Proc. 29th Annu. Int. Conf. IEEE Eng. in Medicine and Biology
  Soc.}, pages 4790--4793, Lyon, France, Aug. 2007.

\bibitem{Carlson2008}
T.~Carlson and Y.~Demiris.
\newblock Human-wheelchair collaboration through prediction of intention and
  adaptive assistance.
\newblock In {\em IEEE Int. Conf. Robot. and Auto.}, pages 1050--4729,
  Pasadena, CA, May 2008.

\bibitem{Carlson2012}
T.~Carlson and Y.~Demiris.
\newblock Collaborative control for a robotic wheelchair: Evaluation of
  performance, attention, and workload.
\newblock {\em IEEE Trans. Syst., Man, and Cybern.}, 42(3), Jun. 2012.

\bibitem{Miyazaki2009}
K.~Miyazaki, M.~Hashimoto, M.~Shimada, and K.~Takahashi.
\newblock Guide following control using laser range sensor for a smart
  wheelchair.
\newblock In {\em ICROS-SICE Int. Joint Conf.}, pages 4613--4616, Fukuoka Int.
  Congr. Center, Japan, Aug. 2009.

\bibitem{Rolland}
A.~Lankenau and T.~Rofer.
\newblock A versatile and safe mobility assistant.
\newblock {\em IEEE Robot. \& Auto. Mag.}, 8(1):29--37, Mar. 2001.

\bibitem{Touati2009}
Y.~Touati, A.~Ali-Ch�rif, and B.~Achili.
\newblock Smart wheelchair design and monitoring via wired and wireless netw.
\newblock In {\em IEEE Symp. Ind. Electron. and Applicat.}, pages 920--925,
  Kuala Lumpur Malaysia, Oct. 2009.

\bibitem{Niitsuma2011}
M.~Niitsuma, T.~Ochi, M.~Yamaguchi, and H.~Hashimoto.
\newblock Design of interaction for simply operating smart electric wheelchair
  in intell. space.
\newblock In {\em 4th Int. Conf. Human System Interactions}, pages 70--75,
  Yokohama, May 2011.

\bibitem{Veno2011}
Y.~Veno, H.~Kitagawa, K.~Kakihara, and K.~Terashima.
\newblock Design and control for collision avoidance of power-assisted
  omni-directional mobile wheelchair system.
\newblock pages 902--907, Kyoto, Dec. 2011.

\bibitem{Viswanathan2011}
P.~Viswanathan, J.~J. Little, A.~K. Mackworth, and A.~Mihailidis.
\newblock Adaptive navigation assistance for visually-impaired wheelchair
  users.
\newblock In {\em Proc. IROS 2011 Workshop New and Emerg. Technol. in Assistive
  Robot.}, San Francisco, CA, USA, Sept. 2011.

\bibitem{Giesbrecht2015}
E.M. Giesbrecht, W.C. Miller, B.T. Jin, I.M. Mitchell, and J.J. Eng.
\newblock Rehab on wheels: A pilot study of tablet-based wheelchair training
  for older adults.
\newblock {\em JMIR Rehabil. and Assistive Technol.}, 2, Apr. 2015.

\bibitem{Urbano2011}
M.~Urbano, J.~Fonseca, U.~Nunes, and H.~Zeilinger.
\newblock Extending a smart wheelchair navigation by stress sensors.
\newblock In {\em IEEE 16th Conf. Emerg. Technol. \& Factory Auto.}, pages
  1--4, Toulouse, Sept. 2011.

\bibitem{sultan2011}
C.~Balaguer, A.~Jardon, C.A. Monje, F.~Bonsignorio, M.F. Stoelen, S.~Martinez,
  and J.G. Victores.
\newblock Sultan: Simultaneous user learning and task execution, and its
  application in assistive robotics.
\newblock In {\em Workshop New and Emerg. Technol. in Assistive Robot.}, San
  Francisco, CA, USA, Sept. 2011.

\bibitem{Poorten2012}
E.~B. Van~Der Poorten, E.~Demeester, E.~Reekmans, and J.~Philips.
\newblock Powered wheelchair navigation assistance through kinematically
  correct environmental haptic feedback.
\newblock In {\em IEEE Int. Conf. Robot. and Auto.}, pages 3706--3712, Saint
  Paul, MN, USA, May 2012.

\bibitem{Tang2012}
R.~Tang, X.~Chen, M.~Hayes, and I.~Palmer.
\newblock Develop. of a navigation system for semi-autonomous operation of
  wheelchairs.
\newblock In {\em IEEE/ASME Int. Conf. Mechatronics and Embedded Syst. and
  Applicat.}, pages 257--262, Suzhou, Jul. 2012.

\bibitem{Carrino2012}
F.~Carrino, J.~Dumoulin, E.~Mugellini, O.~Khaled, and R.~Ingold.
\newblock A self-paced bci system to control an electric wheelchair: evaluation
  of a commercial, low-cost eeg device.
\newblock In {\em ISSNIP Biosignals and Biorobotics Conf.}, pages 1--6, Manaus,
  Jan. 2012.

\bibitem{Nguyen2012}
J.~S. Nguyen.
\newblock {\em A Smart Wheelchair System using a Combination of Stereoscopic
  and Spherical Vision Cameras}.
\newblock PhD thesis, Univ. Technol., Sydney, 2012.

\bibitem{Cockrell2013}
S.~Cockrell, G.~Lee, and W.~Newman.
\newblock Determining navigability of terrain using point cloud data.
\newblock In {\em Proc. IEEE Int. Conf. Rehabil. Robot.}, pages 1--6, Seattle
  Washington, Jun. 2013.

\bibitem{Tavares2013}
J.~Tavares, J.~Barbosa, and C.~Costa.
\newblock A smart wheelchair based on ubiquitous computing.
\newblock In {\em Proc. 6th Int. Conf. Pervasive Technol. Related to Assistive
  Environments}, pages 1--4, New York, NY, USA, May 2013.

\bibitem{Trivedi2013}
A.R. Trivedi, A.K. Singh, S.~T. Digumarti, D.~Fulwani, and S.~Kumar.
\newblock Design and implementation of a smart wheelchair.
\newblock In {\em Proc. Conf. Adv. In Robot.}, pages 1--6, New York, NY, USA,
  2013.

\bibitem{Tyagi2013}
V.~Tyagi, N.~K. Gupta, and P.~K. Tyagi.
\newblock Smart wheelchair using fuzzy inference system.
\newblock In {\em IEEE Global Humanitarian Technol. Conf.: South Asia
  Satellite}, pages 175--180, Trivandrum, Aug. 2013.

\bibitem{Morales2013}
Y.~Morales, N.~Kallakuri, K.~Shinozawa, T.~Miyashita, and Norihiro Hagita.
\newblock Human-comfortable navigation for an auton. robotic wheelchair.
\newblock In {\em IEEE/RSJ Int. Conf. Intell. Robots and Syst.}, pages
  2737--2743, Tokyo, Japan, Nov. 3-7 2013.

\bibitem{Park2013}
S.~Park, T.~T. Ha, J.~Y. Shivajirao, J.~Park, J.~Kim, and M.~Hahn.
\newblock Smart wheelchair control system using cloud-based mobile device.
\newblock In {\em Int. Conf. IT Convergence and Security}, pages 1--3, Macao,
  Dec. 2013.

\bibitem{Fattouh2013}
A.~Fattouh, O.~Horn, and G.~Bourhis.
\newblock Emotional bci control of a smart wheelchair.
\newblock {\em IJCSI Int. J. Comput. Sci. Issues}, 10(3), 2013.

\bibitem{Milenkovic2013}
A.~Milenkovic, M.~Milosevic, and E.~Jovanov.
\newblock Smartphones for smart wheelchairs.
\newblock In {\em IEEE Int. Conf. Body Sensor Netw.}, pages 1--6, Cambridge,
  MA, USA, May 2014.

\bibitem{McMurrough2013}
C.~McMurrough, I.~Ranatunga, A.~Papangelis, D.O. Popa, and F.~Makedon.
\newblock A develop. and evaluation platform for non-tactile power wheelchair
  controls.
\newblock In {\em Proc. 6th Int. Conf. Pervasive Technol. Related to Assistive
  Environments}, pages 1--4, New York, NY, USA, May 2013.

\bibitem{Akash2014}
S.~A. Akash, A.~Menon, A.~Gupta, M.~W. Wakeel, M.~N. Praveen, and P.~Meena.
\newblock A novel strategy for controlling the movement of a smart wheelchair
  using internet of things.
\newblock In {\em IEEE Global Humanitarian Technol. Conf. - South Asia
  Satellite}, pages 154--158, Sept. 2014.

\bibitem{SUGANO2014}
T.~Sugano, Y.~Dan, H.~Okajima, N.~Matsunaga, and Z.~Hu.
\newblock Indoor platoon driving of electric wheelchair with model error
  compensator along wheel track of preceding vehicle.
\newblock In {\em 5th Int. Symp. Adv. Control of Ind. Processes}, Hiroshima,
  Japan, May 2014.

\bibitem{Sato2014}
Y.~Sato, R.~Suzuki, M.~Arai, Y.~Kobayashi, Y.~Kuno, M.~Fukushima, K.~Yamazaki,
  and A.~Yamazaki.
\newblock Multiple robotic wheelchair system able to move with a companion
  using map inform.
\newblock In {\em Proc. ACM/IEEE Int. Conf. Human-robot Interaction}, pages
  286--287, Bielefeld, Germany, Mar. 3�6 2014.

\bibitem{Senthilkumar2014}
S.~Senthilkumar and T.~Shanmugapriya.
\newblock Brain control of a smart wheelchair.
\newblock {\em Int. J. Innovative Research in Comput. and Netw. Eng.}, 2(6),
  2014.

\bibitem{CRIO2014}
A.~Ruiz-Serrano, M.~C. Reyes-Fernandez, R.~Posada-Gomez, A.~Martinez-Sibaja,
  and A.~A. Aguilar-Lasserre.
\newblock Obstacle avoidance embedded system for a smart wheelchair with a
  multimodal navigation interface.
\newblock In {\em 11th Int. Conf. Elect. Eng., Comput. Sci. and Automat.
  Control}, pages 1--6, Campeche, Oct. 2014.

\bibitem{Henderson2014}
M.~Henderson, S.~Kelly, R.~Horne, M.~Gillham, M.~Pepper, and J.~M. Capron.
\newblock Powered wheelchair platform for assistive technology development.
\newblock In {\em 2014 5th Int. Conf. Emerg. Security Technol.}, pages 52--56,
  Alcala de Henares, Sept. 2014.

\bibitem{Cavanini2014}
L.~Cavanini, F.~Benetazzo, A.~Freddiy, S.~Longhi, and A.~Monteriu.
\newblock {SLAM}-based auton. wheelchair navigation system for {AAL} scenarios.
\newblock In {\em IEEE/ASME 10th Int. conference Mechatronic and Embedded Syst.
  and Applicat.}, pages 1--5, Senigallia, Sept. 2014.

\bibitem{Tomari2014}
M.~Tomari, Y.~Kobayashi, and Y.~Kuno.
\newblock {\em The 8th Int. Conf. Robotic, Vision, Signal Process. {\&} Power
  Applicat.: Innovation Excellence Towards Humanistic Technol.}, chapter
  Enhancing Wheelchair's Control Operation of a Severe Impairment User, pages
  65--72.
\newblock Springer Singapore, Singapore, 2014.

\bibitem{Jain2014}
S.~Jain and B.~Argall.
\newblock Automated perception of safe docking locations with alignment inform.
  for assistive wheelchairs.
\newblock In {\em Automated Perception of Safe Docking Locations with Alignment
  Inform. for Assistive Wheelchairs}, pages 4997--03, Chicago IL, USA, Sept.
  2014.

\bibitem{Carrington2014}
P.~Carrington, A.~Hurst, and S.K. Kane.
\newblock Wearables and chairables: Inclusive design of mobile input and output
  techniques for power wheelchair users.
\newblock In {\em Proc. 32nd Annu. ACM Conf. Human Factors in Comput. Syst.},
  pages 3103--3112, New York, NY, USA, 2014.

\bibitem{SRsmartchair}
Smile-Rehab.
\newblock Smart chair.
\newblock {\em https://smilesmart-tech.com/}, Feb. 2017.

\bibitem{Fernandez-Carmona2009}
M.~Fernandez-Carmona, B.~Fernandez-Espejo, J.M. Peula, C.~Urdiales, and
  F.~Sandoval.
\newblock Efficiency based collaborative control modulated by biometrics for
  wheelchair assisted navigation.
\newblock In {\em IEEE 11th Int. Conf. Rehabil. Robot.}, pages 737--742, Kyoto,
  Japan, Jun. 2009.

\bibitem{Felzer2009}
T.~Felzer, B.Strah, R.~Nordmann, and S.~Miglietta.
\newblock Alternative wheelchair control involving intentional muscle
  contractions.
\newblock {\em Int. J. Artificial Intell. Tools}, 18(3):439--465, 2009.

\bibitem{Yokota2011}
S.~Yokota, H.~Hashimoto, D.~Chugo, and J.~She Y~Ohyama.
\newblock Absorption of ambiguous human motion on human body motion interface.
\newblock {\em IEEE Int. Symp. Ind. Electron.}, pages 853--858, Jun. 2011.

\bibitem{Blatt2008}
R.~Blatt, S.~Ceriani, B.D. Seno, Giulio Fontana, M.~Matteuci, and D.~Migliore.
\newblock Brain control of a smart wheelchair.
\newblock {\em Intell. Auton. Syst. 10}, pages 221--228, 2008.

\bibitem{Shen2011}
V.~Shen.
\newblock An intelligent distributed controller for a wheelchair that is
  operated by a brain-computer interface.
\newblock In {\em Dept. of Mechatronics and Bio Med. Eng., Masters thesis},
  Universiti Tunku Abdul Rahman, Jun. 2011.

\bibitem{Carlson2013}
T.~Carlson and J.~Millan.
\newblock Brain�controlled wheelchairs: A robotic architecture.
\newblock {\em IEEE Robot. and Auto. Mag..}, 20(1):65--73, Mar. 2013.

\bibitem{Diez2013}
P.~Diez, S.~Torres Muller, V.~Mut, E.~Laciar, E.~Avila, T.~Bastos-Filho, and
  M.~Sarcinelli-Filho.
\newblock Commanding a robotic wheelchair with a high frequency steady-state
  visual evoked potential based brain-computer interface.
\newblock {\em Med. Eng. \& Physics}, pages 1155--1164, Jan. 2013.

\bibitem{Kannan2013}
R.~Kannan, M.~Athul, A.~Rithun, R.~Manoj.~K. Ajithesh, G.~Tatikonda, and
  U.~Dutt.
\newblock Thought controlled wheelchair using {EEG} acquisition device.
\newblock In {\em 3rd Int. Conf. Advancements in Electron. and Power Eng.},
  Kuala Lumpur, Malaysia, Jan. 2013.

\bibitem{Lopes2013}
A.~Lopes, G.~Pires, and U.~Nunes.
\newblock Assisted navigation for a brain-actuated intell. wheelchair.
\newblock {\em Robot. and Auton. Syst.}, 61(3):245--258, Mar. 2013.

\bibitem{Ahire2014}
N.~Ahire.
\newblock Smart wheelchair by using {EMG} \& {EOG}.
\newblock {\em Int. J. Of Research In Comput. Eng. And Electron..}, 3(5), 2014.

\bibitem{Champaty2014}
B.~Champaty, P.~Dubey, S.~Sahoo, S.S. Ray, and K.~Pal.
\newblock Develop. of wireless {EMG} control system for rehabilitation devices.
\newblock In {\em Int. Conf. Magnetics, Mach. and Drives}, pages 1--4,
  Kottayam, Jul. 2014.

\bibitem{Gandhi2014}
V.~Gandhi, G.~Prasad, D.~Coyle, L.~Behera, and T.~McGinnity.
\newblock {EEG}-based mobile robot control through an adaptive brain-robot
  interface.
\newblock {\em IEEE Trans. Syst., Man, and Cybern.: Syst.}, 44(9):1278--1285,
  Sept. 2014.

\bibitem{Kaufmann2014}
T.~Kaufmann, A.~Herweg, and A.~Kubler.
\newblock Toward brain-computer interface based wheelchair control utilizing
  tactually-evoked event-related potentials.
\newblock {\em J. NeuroEng. \& Rehabil.}, 11(7):1--17, Jan. 2014.

\bibitem{Li2015}
Z.~Li, C.~Yang, S.~Zhao, N.~Wang, and C.-Y. Su.
\newblock Shared control of an intell. wheelchair with dynamic constraints
  using brain-mach. interface.
\newblock {\em Intell. Robot. and Applicat.}, 9245:260--271, Aug. 2015.

\bibitem{Milenkovi2013}
A.~Milenkovi, Mladen M., and E.~Jovanov.
\newblock Smartphones for smart wheelchairs.
\newblock In {\em IEEE Int. Conf. Body Sensor Netw.}, pages 1--6, Cambridge,
  MA, USA, May 2013.

\bibitem{Villalta2006}
H.~Sermeno-Villalta and J.~Spletzer.
\newblock Vision-based control of a smart wheelchair for the automated
  transport and retrieval system {(ATRS)}.
\newblock In {\em Proc. IEEE Int. Conf. Robot. and Auto.}, pages 3423--3428,
  Orlando, FL, USA, May 2006.

\bibitem{Bailey2007}
M.~Bailey, A.~Chanler, B.~Maxwell, M.~Micire, K.~Tsui, and H.~Yanco.
\newblock Develop. of vision-based navigation for a robotic wheelchair.
\newblock In {\em IEEE 10th Int. Conf. Rehabil. Robot.}, pages 951--957,
  Noordwijk, The Netherlands, Jun. 2007.

\bibitem{Ju2009}
J.~Sun Ju, Y.~Shin, and E.~Yi Kim.
\newblock Intell. wheelchair (iw) interface using face and mouth recognition.
\newblock In {\em Proc. 14th Int. Conf. Intell. User Interfaces}, pages
  307--314, Sanibel Island, FL, USA, 2009.

\bibitem{Ahmed2010}
Z.~Ahmed and A.~Shahzad.
\newblock Mobile robot navigation using gaze contingent dynamic interface.
\newblock In {\em Blekinge Inst. of Technol., Masters Thesis in Comput. Sci.},
  Ronneby, Sweden, 2010.

\bibitem{Lee2012}
L.~Chockalingam G.~Chong~Lee, C. Kiong~Loo.
\newblock An integrated approach for head gesture based interface.
\newblock {\em Appl. Soft Comput.}, 12(3):1101--1114, Mar. 2012.

\bibitem{Megalingam2012}
R.~Megalingam and R.~Krishna A.~Thulasi.
\newblock Methods of wheelchair navigation: Novel gesture recognition method.
\newblock {\em Int. J. Appl. Eng. Research}, 7(11):1654--1658, 2012.

\bibitem{Perez2012}
E.~Perez, C.~Soria, O.~Nasisi, T.~Freire Bastos, and V.~Mut.
\newblock Robotic wheelchair controlled through a vision-based interface.
\newblock {\em Robotica}, 30(5):691--708, Sept. 2012.

\bibitem{Kawarazaki2013}
N.~Kawarazaki, D.~Stefanov, and B.~Diaz A.~Israel.
\newblock Toward gesture controlled wheelchair: A proof of concept study.
\newblock In {\em IEEE Int. Conf. Rehabil. Robot.}, pages 1--6, Seattle, WA,
  USA, 2013.

\bibitem{Rivera2013}
L.~Rivera and G.~DeSouza.
\newblock Control of a wheelchair using an adaptive k-means clustering of head
  poses.
\newblock In {\em IEEE Symp. Comput. Intell. in Rehabil. and Assistive
  Technol.}, pages 24--31, Singapore, Apr. 2013.

\bibitem{Kawarazaki2014}
N.~Kawarazaki and M.~Yamaoka.
\newblock Face tracking control system for wheelchair with depth image sensor.
\newblock In {\em 13th Int. Conf. Control, Auto., Robot. \& Vision}.

\bibitem{Ashraf2011}
M.~Ashraf and M.~Ghazali.
\newblock Interaction design for wheelchair using nintendo wiimote controller.
\newblock In {\em Int. Conf. User Sci. and Eng.}, pages 48--53, Shah Alam,
  Selangor, Malaysia, Dec. 2011.

\bibitem{Ashraf2011-2}
M.~Ashraf and M.~Ghazali.
\newblock Augmenting intuitiveness with wheelchair interface using nintendo
  wiimote.
\newblock {\em Int. J. New Comput. Archit. and Their Applicat.}, 1(4):977--990,
  2011.

\bibitem{Gerling2013}
K.M. Gerling, M.R. Kalyn, and R.L. Mandryk.
\newblock {KINECTWheels}: Wheelchair-accessible motion-based game interaction.
\newblock In {\em CHI 2013 Extended Abstracts}, pages 3055--3058, Paris,
  France, 2013.

\bibitem{Sahnoun2006}
M.~Sahnoun and G.~Bourhis.
\newblock Haptic feedback to assist powered wheelchair piloting.
\newblock {\em Assoc. for the Advancement of modeling and simulation techniques
  in entreprises}, 67:53--63, Jan. 2006.

\bibitem{Christensen2011}
Q.~Christensen.
\newblock Three degree of freedom haptic feedback for assisted driving of
  holonomic omnidirectional wheelchairs.
\newblock In {\em Master of Sci. Thesis, Dept. of Mech. Eng.}, Univ. Utah, Aug.
  2011.

\bibitem{Abdelkader2012}
M.~Hadj-Abdelkader, G.~Bourhis, and B.~Cherki.
\newblock Haptic feedback control of a smart wheelchair.
\newblock {\em Appl. Bionics and Biomechanics}, 9(2):181--192, Apr. 2012.

\bibitem{Morre2015}
Y.~Morere, M.A.~Hadj Abdelkader, K.~Cosnuau, and G.~Bourhis G.~Guilmois.
\newblock Haptic control for powered wheelchair driving assistance.
\newblock {\em IRBM}, 36(5):293--304, Oct. 2015.

\bibitem{Serranoa2013}
A.~Ruiz-Serranoa, R.~Posada-Gomez, A.~M. Sibaja, B.~E. Gonzalez-Sanchez
  G.~A.~Rodriguez, and O.~O. Sandoval-Gonzalez.
\newblock Develop. of a dual control system applied to a smart wheelchair,
  using magnetic and speech control.
\newblock {\em Procedia Technol.}, 7:158--165, 2013.

\bibitem{Srivastava2014}
P.~Srivastava, S.~Chatterjee, and R.~Thakur.
\newblock Design and develop. of dual control system appl. to smart wheelchair
  using voice and gesture control.
\newblock {\em Int. J. Research in Elect. \& Electron. Eng.}, 2(2):1--9, 2014.

\bibitem{Reis2015}
L.~Reis, B.~Faria, S.~Vasconcelos, and N.~Lau.
\newblock Multimodal interface for an intell. wheelchair.
\newblock In {\em Inform. in Control, Auto. and Robot., Lecture Notes in Elect.
  Eng. 325}, pages 1--34, Springer Int. Publishing Switzerland, 2015.

\bibitem{Parikh2007}
S.~P. Parikh, V.~Grassi Jr., R.~V. Kumar, and J.~Okamoto Jr.
\newblock Integrating human inputs with auton. behaviors on an intell.
  wheelchair platform.
\newblock {\em IEEE Intell. Syst.}, 22(2):33--41, 2007.

\bibitem{Faria2014}
B.~M. Faria, L.~P. Reis, and N.~Lau.
\newblock A survey on intelligent wheelchair prototypes and simulators.
\newblock {\em New Perspectives in Inform. Syst. and Technol.}, 1:545--557,
  2014.

\bibitem{Nguyen2013}
J.S. Nguyen, S.W. Su, and H.T. Nguyen.
\newblock Experimental study on a smart wheelchair system using a combination
  of stereoscopic and spherical vision.
\newblock In {\em 35th Annu. Int. Conf. IEEE Eng. in Medicine and Biology
  Soc.}, pages 4597--4600, Osaka, Japan, Jul. 2013.

\bibitem{IMU}
Inertial Measurement~Unit (\textsc{IMU}).
\newblock http://www.microstrain.com/.

\bibitem{Benavidez2011}
P.~Benavidez and M.~Jamshidi.
\newblock Mobile robot navigation and target tracking system.
\newblock {\em 6th Int. Conf. System of Syst. Eng.}, pages 299--304, Jun. 2011.

\bibitem{Kulp2012}
W.~Kulp.
\newblock Robotic person-following in cluttered environments.
\newblock {\em Case Western Reserve Univ. EECS Dept. Masters Thesis}, 2012.

\bibitem{Fallon2012}
M.~Fallon, H.~Johannsson, and J.~Leonard.
\newblock Efficient scene simulation for robust monte carlo localization using
  an {RGB-D} camera.
\newblock {\em IEEE Int. Conf. Robot. and Auto.}, pages 1663--1670, May 2012.

\bibitem{NavChair}
S.~P. Levine, D.~A. Bell, L.~A. Jaros, R.~C. Simpson, Y.~Koren, and
  J.~Borenstein.
\newblock The navchair assistive wheelchair navigation system.
\newblock {\em IEEE Trans. Rehabil. Eng.}, 7(4):443 --451, Dec. 1999.

\bibitem{Touati2012}
Y.~Touati and A.~Ali-Cherif.
\newblock Smart powered wheelchair platform design and control for people with
  severe disabilities.
\newblock {\em Software Eng.}, 2(3):49--56, 2012.

\bibitem{Kobayashi2012}
Y.~Kobayashi, Y.~Kinpara, E.~Takano, Y.~Kuno, K.~Yamazaki, and A.~Yamazaki.
\newblock Robotic wheelchair moving with caregiver collaboratively.
\newblock {\em Adv. Intell. Comput. Theories and Applicat.. With Aspects of
  Articial Intell.}, 6839:523--532, 2012.

\bibitem{Murakami2014}
R.~Murakami, Y.~Morales, S.~Satake, T.~Kanda, and H.~Ishiguro.
\newblock Destination unknown: Walking side-by-side without knowing the goal.
\newblock pages 471--478, 2014.

\bibitem{Suzuki2014}
R.~Suzuki, T.~Yamada, M.~Arai, Y.~Sato, Y.~Kobayashi, and Y.~Kuno.
\newblock Multiple robotic wheelchair system considering group commun.
\newblock In {\em Adv. in Visual Comput.: 10th Int. Symp.}, Lecture Notes in
  Comput. Sci., pages 805--814. Las Vegas, NV, USA, Dec. 2014.

\bibitem{JPLGrids}
H.~P. Moravec.
\newblock Sensor fusion in certainty grids for mobile robots.
\newblock {\em Sensor Devices and Syst. for Robot.}, 52:253--276, 1989.

\bibitem{Wei2012_IRA}
W.~Zhixuan, W.~Chen, and J.~Wang.
\newblock 3d semantic map-based shared control for smart wheelchair.
\newblock In Chun-Yi Su, Subhash Rakheja, and Honghai Liu, editors, {\em
  Intell. Robot. and Applicat.}, volume 7507 of {\em Lecture Notes in Comput.
  Sci.}, pages 41--51. Springer Berlin Heidelberg, 2012.

\bibitem{Yuki2015_AR}
Yuki Uratsuji, Kentaro Takemura, Jun Takamatsu, and Tsukasa Ogasawara.
\newblock Mobility assistance system for an electric wheelchair using annotated
  maps.
\newblock {\em Adv. Robot.}, 29(7):481--491, 2015.

\bibitem{Yang2011}
L.~Yang, D.~Zhao, X.~Wu, H.~Li, and J.~Zhai.
\newblock An improved prewitt algorithm for edge detection based on noised
  image.
\newblock {\em 2011 4th Int. Congr. on Image and Signal Process.},
  3:1197--1200, Oct. 2011.

\bibitem{Rofer2009}
T.~Rofer, C.~Mandel, and T.~Laue.
\newblock Controlling an automated wheelchair via joystick/head-joystick
  supported by smart driving assistance.
\newblock In {\em 2009 IEEE Int. Conf. Rehabil. Robot.}, pages 743--748, Kyoto
  Japan, Jun. 2009.

\bibitem{Ren2012}
Y.~Ren, W.~Zou, H.~Fan, A.~Ye, K.~Yuan, and Y.~Ma.
\newblock A docking control method in narrow space for intelligent wheelchair.
\newblock In {\em 2012 IEEE Int. Conf. Mechatronics and Auto.}, pages
  1615--1620, Chengdu, China, Aug. 2012.

\bibitem{AVNguyen2013}
A.V. Nguyen, L.B. Nguyen, S.~Su, and H.T. Nguyen.
\newblock The advancement of an obstacle avoidance bayesian neural network for
  an intell. wheelchair.
\newblock In {\em 2013 35th Annu. Int. Conf. IEEE Eng. in Medicine and Biology
  Soc.}, pages 3642--3645, Osaka Japan, Jul. 2013.

\bibitem{Takano2012}
E.~Takano, Y.~Kobayashi, and Y.~Kuno.
\newblock Multiple robotic wheelchair system based on the observation of
  circumstance.
\newblock {\em 18th Korea-Japan Joint Workshop Frontiers of Comput. Vision},
  pages 222--226, 2012.

\bibitem{Iida1991}
S.~Iida and S.~Yuta.
\newblock Vehicle command system and trajectory control for autonomous mobile
  robots.
\newblock {\em Proc. IEEE/RSJ Int. Workshop Intell. Robots and Syst., Intell.
  for Mech. Syst.}, 1:212--217, 1991.

\bibitem{Li2011}
Q.~Li, W.~Chen, and J.~Wang.
\newblock Dynamic shared control for human-wheelchair cooperation.
\newblock In {\em 2011 IEEE Int. Conf. Robot. and Auto.}, pages 4278--4283, May
  2011.

\bibitem{Ohno2004}
K.~Ohno, T.~Tsubouchi, B.~Shigematsu, and S.~Yuta.
\newblock Differential \textsc{GPS} and odometry-based outdoor navigation of a
  mobile robot.
\newblock {\em Adv. Robot.}, 18(6):611--635, 2004.

\bibitem{North2012}
E.~North, J.~Georgy, U.~Iqbal, M.~Tarbochi, and A.~Noureldin.
\newblock Improved inertial/odometry/\textsc{GPS} positioning of wheeled robots
  even in \textsc{GPS}-denied environments.
\newblock {\em Global Navigation Satellite Syst.: Signal, Theory and
  Applicat.}, 11:257--278, Feb. 2012.

\bibitem{La2013TMECH}
H.~M. La, R.~S. Lim, B.~B. Basily, N.~Gucunski, J.~Yi, A.~Maher, F.~A. Romero,
  and H.~Parvardeh.
\newblock Mechatronic and control systems design for an autonomous robotic
  system for high-efficiency bridge deck inspection and evaluation.
\newblock {\em IEEE Trans. Mechatronics}, 18(6):1655--1664, Apr. 2014.

\bibitem{La2014IROS}
H.~M. La, N.~Gucunski, S.H. Kee, J.~Yi, T.~Senlet, and L.~Nguyen.
\newblock Auton. robotic system for bridge deck data collection and anal.
\newblock {\em 2014 IEEE/RSJ Int. Conf. Intell. Robots and Syst.}, pages
  1950--1955, Sept. 2014.

\bibitem{LaCASE13_RABIT}
H.~M. La, R.~S. Lim, B.~B. Basily, N.~Gucunski, J.~Yi, A.~Maher, F.~A. Romero,
  and H.~Parvardeh.
\newblock Auton. robotic system for high-efficiency non-destructive bridge deck
  inspection and evaluation.
\newblock {\em 2013 IEEE Int. Conf. Auto. Sci. and Eng.}, pages 1053--1058,
  2013.

\bibitem{bumblebee}
PointGrey.
\newblock Bumblebee xb3 1394b.
\newblock {\em www.ptgrey.com}, Feb. 2015.

\bibitem{ladybug}
PointGrey.
\newblock Bgr ladybug2.
\newblock {\em www.ptgrey.com}, Feb. 2015.

\bibitem{Padir2014}
T.~Padir.
\newblock Towards personalized smart wheelchairs: Lessons learned from
  discovery interviews.
\newblock In {\em 2015 37th Annu. Int. Conf. IEEE Eng. in Medicine and Biology
  Soc.}, Milan, 2014.

\bibitem{RESNA2011}
{RESNA}~Board of~Directors.
\newblock {RESNA} wheelchair service provision guide.
\newblock In {\em Rehabil. Eng. \& Assistive Technol. Soc. of North America},
  Jan. 2011.

\bibitem{Gerling2014}
K.~M. Gerling, R.~L. Mandryk, M.~V. Birk, M.~Miller, and R.~Orji.
\newblock The effects of embodied persuasive games on player attitudes toward
  people using wheelchairs.
\newblock In {\em Proc. 32nd Annu. ACM Conf. Human Factors in Comput. Syst.},
  pages 3413--3422, Toronto, Ontario, Canada, 2014.

\bibitem{Narayanan2014}
V.~K. Narayanan, F.~Pasteau, M.~Babel, and F.~Chaumette.
\newblock Visual servoing for autonomous doorway passing in a wheelchair using
  a single doorpost.
\newblock In {\em IEEE/RSJ IROS Workshop Assistance and Service Robot. in a
  Human Environment}, Chicago, IL, USA, 2014.

\bibitem{Mitchell2014}
I.~M. Mitchell, P.~Viswanathan, B.~Adhikari, E.~Rothfels, and A.~K. Mackworth.
\newblock Shared control policies for safe wheelchair navigation of elderly
  adults with cognitive and mobility impairments: Designing a wizard of oz
  study.
\newblock In {\em 2014 Amer. Control Conf.}, pages 4087--4094, Portland, OR,
  USA, Jun. 2014.

\bibitem{Fehr:2000}
L.~Fehr.
\newblock {Adequacy of power wheelchair control interfaces for persons with
  severe disabilities: a clinical survey}.
\newblock {\em J. Rehabil. Research and Develop.}, 37(3):353--360, May/Jun.
  2000.

\bibitem{Carlson2013_RAM}
T.~Carlson and J.~D.R.~Millan.
\newblock Brain-controlled wheelchairs: A robotic architecture.
\newblock {\em IEEE Robot. Auto. Mag.}, 20:65--73, Mar. 2013.

\bibitem{Gunachandra2014}
Gunachandra, S.~Chrisander, A.~Widyotriatmo, and Suprijanto.
\newblock Wall following control for the application of a brain-controlled
  wheelchair.
\newblock In {\em 2014 IEEE Int. Conf. Intell. Auton. Agents, Netw. and Syst.},
  pages 36--41, Aug. 2014.

\bibitem{Jang2014_IROS}
G.~Jang and Y.~Choi.
\newblock {EMG}-based continuous control method for electric wheelchair.
\newblock In {\em 2014 IEEE/RSJ Int. Conf. Intell. Robots and Syst.}, pages
  3549--3554, Sept. 2014.

\bibitem{Widyotriatmo2015_ASCC}
A.~Widyotriatmo, Suprijanto, and S.~Andronicus.
\newblock A collaborative control of brain computer interface and robotic
  wheelchair.
\newblock In {\em 2015 10th Asian Control Conf.}, pages 1--6, May 2015.

\bibitem{Hancock2011}
P.~A. Hancock, D.~R. Billings, K.~E. Schaefer, J.~Y.~C. Chen, E.~J. de~Visser,
  and R.~Parasuraman.
\newblock A meta-anal. of factors affecting trust in human-robot interaction.
\newblock {\em Human Factors}, 53(5):517--527, Oct. 2011.

\bibitem{Riek2014}
L.~D. Riek and D.~Howard.
\newblock A code of ethics for the human-robot interaction profession.
\newblock In {\em Proc. We Robot}, Apr. 2014.

\bibitem{Alrajhi2014}
W.~Alrajhi, M.~Hosny, A.~Al-Wabil, and A.~Alabdulkarim.
\newblock Human factors in the design of bci-controlled wheelchairs.
\newblock {\em Human-Comput. Interaction. Adv. Interaction Modalities and
  Tech.}, 8511:513--522, 2014.

\bibitem{CanWheel}
CanWheel publications.
\newblock Canwheel.
\newblock {\em www.canwheel.ca}, Jul. 2015.

\bibitem{Gironimo2013}
G.D. Gironimo, G.~Matrone, A.~Tarallo, and M.~Trotta.
\newblock A virtual reality approach for usability assessment: case study on a
  wheelchair-mounted robot manipulator.
\newblock {\em Eng. with Comput.}, 29:359--373, 2013.

\bibitem{Adhikari2014}
B.~Adhikari.
\newblock A single subject participatory action design method for powered
  wheelchairs providing automated back-in parking assistance to cognitively
  impaired older adults: A pilot study.
\newblock In {\em Master of Sci. Thesis, Comput. Sci., Univ. British Columbia},
  2014.

\bibitem{Jipp2012}
M.~Jipp.
\newblock Individual differences and their impact on the safety and the
  efficiency of human-wheelchair syst..
\newblock {\em Human Factors}, 54(6):1075--1086, Dec. 2012.

\bibitem{KANKI2014}
Y.~Kanki, N.~Kuwahara, and K.~Morimoto.
\newblock An evaluation of physical strains while driving an electric
  wheelchair.
\newblock In {\em 2014 IIAI 3rd Int. Conf. Adv. Appl. Inform.}, pages 863--866,
  Kitakyushu, Japan, Sept. 2014.

\bibitem{Giesbrecht2014}
E.~M. Giesbrecht, W.~C. Miller, I.~M. Mitchell, and R.~L. Woodgate.
\newblock Develop. of a wheelchair skills home program for older adults using a
  participatory action design approach.
\newblock {\em BioMed Research Int.}, 2014(172434):1--13, 2014.

\bibitem{Shiomi2014}
M.~Shiomi, T.~Iio, K.~Kamei, C.~Sharma, and N.~Hagita.
\newblock User-friendly auton. wheelchair for elderly care using ubiquitous
  network robot platform.
\newblock In {\em Proc. 2nd Int. Conf. Human-agent Interaction}, pages 17--22,
  Tsukuba, Japan, Oct. 2014.

\bibitem{Jensen2014}
L.~Jensen.
\newblock User perspectives on assistive technology: a qualitative analysis of
  55 letters from citizens applying for assistive technology.
\newblock {\em World Federation of Occupat.al Therapists Bulletin},
  69(1):42--45, May 2014.

\bibitem{EZlitecruiser}
{https://www.ezlitecruiser.com}.

\bibitem{KDsmartchair}
{http://medmartonline.com/kd-smartchair-kd-healthcare}.

\bibitem{smartdrive}
{http://medmartonline.com/max-mobility-smartdrive-mx2}.

\bibitem{Greer2012}
N.~Greer, M.~Brasure, and T.~J. Wilt.
\newblock Wheeled mobility (wheelchair) service delivery: scope of the
  evidence.
\newblock {\em Ann. Internal Medicine}, 156(2):141--6, Jan. 2012.

\bibitem{hhs}
U.S~Dept. of~Health and Human Services.~Office for Human Research~Protections.
\newblock Code of federal regulations part 46 - protection of human subjects.
\newblock {\em
  http://www.hhs.gov/ohrp/regulations-and-policy/regulations/45-cfr-46/index.html}.

\bibitem{Bourhis2001}
Bourhis G, Horn O, Habert O, and Pruski A.
\newblock An autonomous vehicle for people with motor disabilities.
\newblock {\em IEEE Robot. \& Auto. Mag.}, 8(1):20--28, Mar. 2001.

\bibitem{Pretz2013}
Pretz K.
\newblock Building smarter wheelchairs. making life easier for people can't
  walk.
\newblock {\em The Inst.-IEEE news source}, Jun. 7, 2013.

\bibitem{Rocha2014}
B.~M. Faria, L.~P. Reis, and N.~Lau.
\newblock {\em New Perspectives in Inform. Syst. and Technol., Volume 1},
  volume 275 of {\em Adv. in Intell. Syst. and Comput.}, chapter A Survey on
  Intell. Wheelchair Prototypes and Simulators, pages 545--557.
\newblock Springer Int. Publishing, 2014.

\bibitem{La2011ICRA}
R.~S. Lim, H.~M. La, Z.~Shan, and W.~Sheng.
\newblock Developing a crack inspection robot for bridge maintenance.
\newblock {\em 2011 IEEE Int. Conf. Robot. and Auto.}, pages 6288--6293, May
  2011.

\bibitem{La2014TASE}
R.~S. Lim, H.~M. La, and W.~Sheng.
\newblock A robotic crack inspection and mapping system for bridge deck
  maintenance.
\newblock {\em IEEE Trans. Auto. Sci. and Eng.}, 11(2):367--378, Apr. 2014.

\bibitem{Sutton1998}
R.~S. Sutton and A.~G. Barto.
\newblock {\em Reinforcement Learning: An Introduction}.
\newblock The MIT Press, Cambridge, Massachusetts, 1998.

\bibitem{Reis2014}
L.~P. Reis, B.~M. Faria, S.~Vasconcelos, and N.~Lau.
\newblock Invited paper: Multimodal interface for an intell. wheelchair.
\newblock volume 325 of {\em Lecture Notes in Elect. Eng.}, pages 1--34.
  Springer Int. Publishing, 2015.

\bibitem{Emotiv}
Emotiv.
\newblock Brain activity tracker.
\newblock {\em Emotive sensors, Available at: http://emotiv.com/}, Mar. 2014.

\bibitem{Peters2013}
J.-P. Peters and V.~Collin.
\newblock Vision systems and the lives of people with disabilities.
\newblock In {\em 2013 IEEE Appl. Imagery Pattern Recognition Workshop: Sensing
  for Control and Augmentation}, pages 1--6, Oct. 2013.

\bibitem{Drumwright2004}
E.~Drumwright, O.~C. Jenkins, and M.~J. Mataric.
\newblock Exemplar-based primitives for humanoid movement classification and
  control.
\newblock In {\em Proc. 2004 IEEE Int. Conf. Robot. and Auto.}, volume~1, pages
  140--145, Apr. 2004.

\bibitem{Narayanan2007}
D.~Wang and S.~Narayanan.
\newblock An acoustic measure for word prominence in spontaneous speech.
\newblock {\em IEEE Trans. Speech, Audio and Language Process.},
  2(15):690--701, 2007.

\bibitem{Obaidat2014}
R.~Covarrubias, A.~Covarrubias, C.~Vilda, I.~M. Diego, and E.~Cabello.
\newblock Controlling a wheelchair through head movement through artificial
  vision and using speech recognition.
\newblock In {\em E-Bus. and Telecommun.}, pages 242--254. 2014.

\bibitem{Rizzolatti1998}
G.~Rizzolatti and M.~A. Arbib.
\newblock Language within our grasp.
\newblock {\em Trends in Neurosciences}, 5(21):188--194, 1998.

\end{thebibliography}
\vspace{-30pt}
\begin{IEEEbiography}[{\includegraphics[width=1in,keepaspectratio]{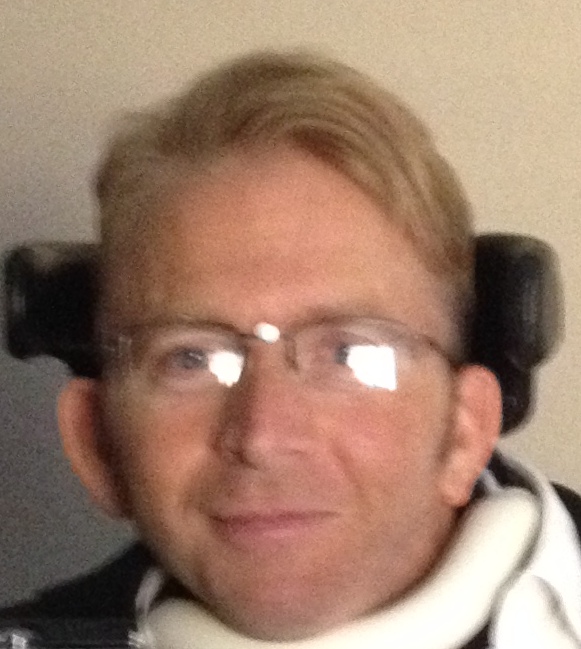}}]{Jesse Leaman} received his B.S. in Astronomy from the University of Maryland, College Park, MD, USA, in 2002. He earned his M.S. and Ph.D. in Astrophysics from the University of California, Berkeley, CA, USA, in 2004, and 2008, respectively.  He was a Post Doctoral research fellow of the NASA's Ames Research Center, Moffett Field, CA, USA, from  2008-2010. Since 2013 he has been a Councilman on the Nevada Assistive Technology Council, Reno, NV, USA. Dr. Leaman currently is a Post Doctoral research fellow of the Advanced Robotics and Automation (ARA) Lab, Department of Computer Science and Engineering, University of Nevada, Reno, NV, USA. 
\vspace{-30pt}
\end{IEEEbiography}
\begin{IEEEbiography}[{\includegraphics[width=1in,height=1.25in,clip,keepaspectratio]{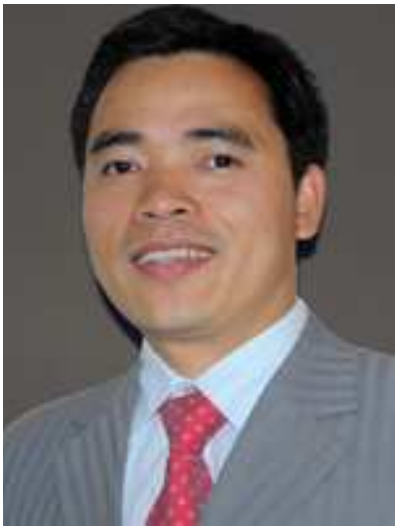}}]{Hung M. La} 
(IEEE SM'2014, M'2009) received his B.S. and M.S. degrees in Electrical Engineering from Thai Nguyen University of Technology, Thai Nguyen, Vietnam, in 2001 and 2003, respectively, and his Ph.D. degree in Electrical and Computer Engineering from Oklahoma State University, Stillwater, OK, USA, in 2011.
He is the Director of the Advanced Robotics and Automation (ARA) Lab, and Assistant Professor of the Department of Computer Science and Engineering, University of Nevada, Reno, NV, USA.  From 2011 to 2014, he was a Post Doctoral research fellow and then a Research Faculty Member at the Center for Advanced Infrastructure and Transportation, Rutgers University, Piscataway, NJ, USA.  Dr. La is an Associate Editor of the IEEE Transactions on Human-Machine Systems,  and Guest Editor of International Journal of Robust and Nonlinear Control. \
\vspace{-30pt}
\end{IEEEbiography}
\end{document}